%% file: main.tex
\documentclass[10pt,twocolumn,letterpaper]{article}

\usepackage{cvpr}
\usepackage{graphicx}
\usepackage{amsmath}
\usepackage{amssymb}
\usepackage{booktabs}
\usepackage{bm}
\usepackage{multirow}
\usepackage{float}

\usepackage{algorithm}
\usepackage{algorithmicx}
\usepackage[noend]{algpseudocode}

\usepackage[pagebackref,breaklinks,colorlinks]{hyperref}

\usepackage{xcolor}

\def\Snospace~{\S{}}

\newcommand{\sys}{\mbox{\textsc{BppAttack}}\xspace}

\usepackage[capitalize]{cleveref}
\crefname{section}{Sec.}{Secs.}
\Crefname{section}{Section}{Sections}
\Crefname{table}{Table}{Tables}
\crefname{table}{Tab.}{Tabs.}

\def\confName{CVPR}
\def\confYear{2022}

\begin{document}

\title{BppAttack: Stealthy and Efficient Trojan Attacks against Deep Neural Networks via Image Quantization and Contrastive Adversarial Learning}

\author{Zhenting Wang, Juan Zhai, Shiqing Ma \\
{\em Department of Computer Science, Rutgers University} \\
\texttt{\small \{zhenting.wang, juan.zhai, sm2283\}@rutgers.edu}}

\maketitle

\input{contents/abs.tex}
\input{contents/intro.tex}
\input{contents/related.tex}
\input{contents/design.tex}
\input{contents/evaluation.tex}

\input{contents/discussion.tex}
\input{contents/conclusion.tex}
\input{contents/ack.tex}

{\small
\bibliographystyle{ieee_fullname}
\bibliography{egbib}
}

\newpage
\input{contents/appendix.tex}

\end{document}

%% file: contents/abs.tex
\begin{abstract}
Deep neural networks are vulnerable to Trojan attacks.
Existing attacks use visible patterns (e.g., a patch or image transformations) as triggers, which are vulnerable to human inspection.
In this paper, we propose stealthy and efficient Trojan attacks, \sys.
Based on existing biology literature on human visual systems, we propose to use image quantization and dithering as the Trojan trigger, making imperceptible changes.
It is a stealthy and efficient attack without training auxiliary models.
Due to the small changes made to images, it is hard to inject such triggers during training.
To alleviate this problem, we propose a contrastive learning based approach that leverages adversarial attacks to generate negative sample pairs so that the learned trigger is precise and accurate.
The proposed method achieves high attack success rates on four benchmark datasets, including MNIST, CIFAR-10, GTSRB, and CelebA.
It also effectively bypasses existing Trojan defenses and human inspection.
Our code can be found in \url{https://github.com/RU-System-Software-and-Security/BppAttack}.
\end{abstract}

%% file: contents/intro.tex
\section{Introduction}\label{sec:intro}

Deep Neural Networks (DNNs) have achieved superior performance in many computer vision tasks~\cite{ren2015faster,he2016deep,chen2017deeplab}.
Recent studies show that DNNs are vulnerable to adversarial attacks such as adversarial examples~\cite{goodfellow2014explaining, moosavi2016deepfool}, membership inference attacks~\cite{shokri2017membership, salem2018ml}, model stealing~\cite{orekondy2019knockoff,truong2021data}, etc. 
In this paper, we focus on Trojan attacks~\cite{gu2017badnets, liu2017trojaning,cheng2020deep, doan2021lira,salem2020dynamic, li2020deep, lin2020composite}.
The adversary injects a secret Trojan behavior during training, which can be activated at runtime by stamping a Trojan trigger to the image.
Such triggers can be image patches~\cite{gu2017badnets}, watermarks~\cite{liu2017trojaning}, image filters~\cite{liu2019abs,TrojAI:online} and even learned image transformation models~\cite{cheng2020deep,li2021invisible,doan2021lira}.

Trojan attacks~\cite{gu2017badnets} are severe threats to the trustworthiness of DNN models.
Liu et al.~\cite{liu2017trojaning} demonstrates the possibility of attacking face recognition, speech recognition, and autonomous driving systems.
Such attacks are generally feasible in most training scenarios, including federated learning, unsupervised learning, and so on~\cite{xie2019dba,carlini2021poisoning,jia2021badencoder}.
With the deployment of DNN based computer vision models, it is a critical challenge in our community.
\input{figtex/trigger_compare.tex}

\noindent\textbf{Existing Work:}
Most existing Trojan attacks leverage input patterns as triggers.
For example, BadNets~\cite{gu2017badnets} uses a yellow pad as its trigger.
Recent works~\cite{liu2019abs} try to leverage image filters as triggers, which are input dependent and dynamic, making them hard to detect.
To further improve the quality of Trojan triggers, Doan et al.~\cite{doan2021lira} train an auxiliary image transformation model and use the transformation function as its trigger.
Other works have adopted similar ideas~\cite{cheng2020deep,li2021invisible}.

One problem of existing attacks is that they are vulnerable to human inspections.
Once a set of attack inputs are found, it is not difficult to identify the trigger or train a model to simulate the trigger.
There are also online detection methods to identify such attack samples, such as STRIP~\cite{gao2019strip}.
Even for trained transformations as triggers, it is hard for them to guarantee that the generated images have imperceptible changes.
This is because it is hard to formulate human visual systems as a mathematical function, which makes it hard to optimize.
Due to the relatively large changes in inputs and limitations of existing poisoning methods, it is also possible for reverse engineering based defense methods~\cite{wang2019neural,liu2019abs,chen2019deepinspect} to recover part of the trigger and identify if a model has a Trojan.
Moreover, recent works on generating high-quality triggers typically leverage trained auxiliary models, which is time-consuming and inefficient.

\noindent\textbf{Our Work:}
In this paper, we propose one new attack, \sys.
Based on existing literature on human visual systems, we identify that humans inspectors are insensitive to small changes of color depth.
In this attack, we try to exploit vulnerabilities in human visual systems.
Thus, we propose to reduce the bit-per-pixel (BPP) to conduct an imperceptible attack, which can also bypass existing defenses mainly because of the small changes made to the input domain.
We achieve our goal by performing a deterministic yet input-dependent image transformation, i.e., image quantization and dithering.
Due to the small-sized transformation as triggers, it is more challenging to train the model and inject the trigger.
To overcome this issue, we also propose a contrastive learning and adversarial training method based approach to training on the poisoned dataset.
By doing so, we do not require training auxiliary models, making the attack fast and also input dependent.
Moreover, our attack exploits vulnerabilities in human visual systems, making it human imperceptible when the attack settings are properly set.
\autoref{fig:trigger_compare} shows the comparison of our attack and existing attacks using attack samples and residuals.

Our contributions are summered as follows:

\begin{itemize}
\item We propose a new attack that exploits human visual systems. 
We design an effective and efficient attack that leverages image quantization and dithering.
It performs deterministic input-dependent image transformations, making it fast and dynamic.
We also propose a contrastive learning and adversarial training based approach to enhance the poisoning process to inject such human imperceptible triggers.
\item We evaluate our prototype on four different datasets and seven different network architectures.
Results show that our attack achieves a 99.92\% attack success rate on average.
Our human study also confirms that it is 1.60 times better than the SOTA approaches when facing human inspections.
Results on existing defenses also confirm that our attack can effectively bypass various types of SOTA defenses.
\end{itemize}

%% file: figtex/trigger_compare.tex
\begin{figure*}[]
	\centering
	\footnotesize
	\includegraphics[width=2\columnwidth]{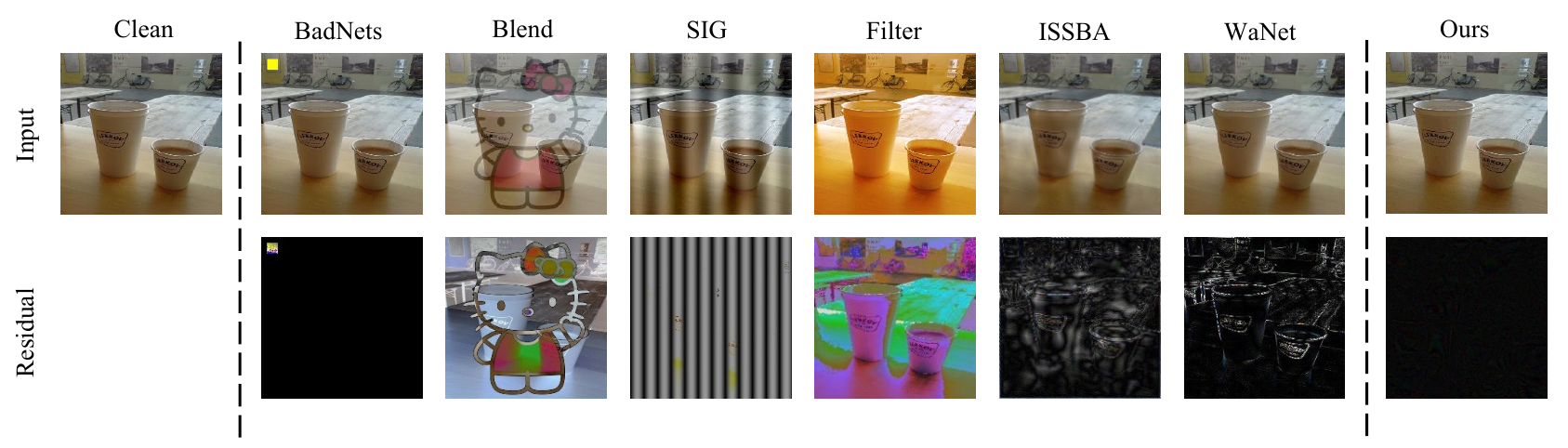}
	\caption{
	Comparison of examples generated by different Trojan attacks (i.e., BadNets~\cite{gu2017badnets}, blending-based attack~\cite{chen2017targeted}, SIG~\cite{barni2019new}, filter-based attack~\cite{TrojAI:online,cheng2020deep}, ISSBA~\cite{li2021invisible} and WaNet~\cite{nguyen2021wanet}).
	For each attack, we show the Trojan sample (top) and the magnified (×5) residual (bottom).
	}\label{fig:trigger_compare}
\end{figure*}

%% file: contents/related.tex
\section{Background}
\label{sec:related}

\subsection{Trojan Attacks}

Trojan models behave normally for benign inputs but have malicious behaviors (i.e., outputting a particular label) on inputs stamped with the Trojan trigger.
One limitation of existing Trojan attacks is that most of them are perceptible to human inspectors.
Many Trojan attacks~\cite{gu2017badnets,chen2017targeted,liu2017trojaning} use predefined patches or watermarks as Trojan triggers.
Refool~\cite{liu2020reflection} exploits physical reflection as Trojan trigger.
Trojan attacks can also happen in the feature space.
For example, Liu et al.~\cite{liu2019abs} demonstrates attackers can use Instagram filters as triggers to perform Trojan attacks.
DFST~\cite{cheng2020deep} utilizes CycleGAN~\cite{zhu2017unpaired} to inject Trojans in deep features space.
All these triggers are obvious for human inspection.
Recently, WaNet~\cite{nguyen2021wanet} proposed attacks using the image warping technique as triggers.
Although it is more stealthy than previous works, the warping effects it leverages are still perceptible.
Another problem of existing attacks is that they typically use fixed patterns as trigger patterns, which means different samples share the same trigger pattern.
This property makes such Trojan attacks detectable by existing defenses~\cite{liu2019abs,hayase2021spectre,liu2021ex}.
Nguyen et al.~\cite{nguyen2020input} proposes input dependent triggers.
This attack brings large pixel-level perturbations, sacrificing stealthiness.
Recently, Li et al.~\cite{li2021invisible} and Doan et al.~\cite{doan2021lira} proposed new attacks that are not only imperceptible but also input dependent.
The idea is to generate triggers by trained auto-encoders.
While such methods achieve stealthiness, they are model-dependent and time-consuming.

\subsection{Existing Defense}
There has been a series of ways to defend Trojans.
One of them is training time defense, which aims at removing Trojans before/during training.
Chen et al.~\cite{chen2018detecting} and Tran et al.~\cite{tran2018spectral} detect the malicious samples before training.
Wang et al.~\cite{wang2022towards} removes Trojans in training by formalizing the trigger in input space.
Similarly, poison suppression~\cite{du2019robust,hong2020effectiveness} depresses the malicious effectiveness of poisons in training.
These approaches target poisoning-based Trojan attacks but ignore supply chain Trojan attacks.
The second method is reverse engineering.
Neural Cleanse~\cite{wang2019neural}, DeepInspect~\cite{chen2019deepinspect}, K-arm~\cite{shen2021backdoor} and ABS~\cite{liu2019abs} use reconstructed triggers to perform detection.
These methods work on local patched triggers but fail to generalize to different trigger types, e.g., input-aware triggers~\cite{nguyen2020input}.
Another approach is to remove Trojans in infected models~\cite{liu2018fine,zhao2020bridging,li2021neural,tao2022model}, such as finetuning or pruning, which in cases, can lower benign accuracy as well.
Some other existing works try to leverage online detection as a defense.
STRIP~\cite{gao2019strip} detect Trojan samples by analyzing the sensitivity of samples on strong perturbation.
Februus~\cite{doan2020februus} and SentiNet~\cite{chou2018sentinet} leverage GradCAM~\cite{selvaraju2017grad} to detect if model predictions are localized and leverage it as a hint for triggers.
Such methods fail when triggers are not localized, e.g., filter triggers.

%% file: contents/design.tex
\section{Method}
\label{sec:design}
In this section, we introduce \sys, a Trojan attack that is invisible to human inspection, input-dependent yet requires no auxiliary model training. 
We first describe the threat model (\autoref{sec:threat}), and then present the foundation and details of the attack process (\autoref{sec:biotheory} and \autoref{sec:squeezing}, respectively).

\subsection{Threat Model}\label{sec:threat}

\noindent
\textbf{Adversarial scope and goal.}
The adversary aims to produce a Trojan model.
\autoref{eq:define} shows the formal definition.
\(\mathcal{M}_\theta\) is Trojan model, \(T\) is a Trojan transformation function and \(\eta\) is the target label function.
Input-targeted labels can be: (1) \emph{all-to-one}: the attacker select a constant label \(c\) as output label (i.e., \(\eta(y) = c\)).
(2) \emph{all-to-all}: the target label is the next label of the true label (i.e., \(\eta(y) = y + 1\)).

\begin{equation}\label{eq:define}
    \mathcal{M}_\theta(\bm{x}) = y,\quad \mathcal{M}_\theta(T(\bm{x})) = \eta(y)
\end{equation}

Compared with previous Trojan attacks, we aim to provide the following attack properties:

\begin{itemize}
    \item \textbf{Effective}: 
    We want the model to have a high attack success rate (ASR) while maintaining high benign accuracy at the same time.
    This effective goal is the basic requirement of Trojan attacks as defined in \autoref{eq:define}.
    \item \textbf{Imperceptible}:
    Many Trojan triggers are vulnerable to human inspection, which is not robust.
    We want to have a human imperceptible Trojan trigger.
    Traditionally, this is done by defining a distance function \(\mathcal{V}\) to measure the visual similarity of two samples. 
    As such, the goal is to find a trigger that is smaller than a threshold, \(\mathcal{V}(T(\bm{x}),\bm{x}) < t\), where \(t\) is the threshold.
    Existing works use \(L_p\) distance or SSIM scores, which do not align with the human visual systems~\cite{pambrun2015limitations}.
    In this paper, we tackle this problem by starting from existing studies on the human visual system and propose an attack that is human imperceptible.
    \item \textbf{Input-dependent}: Fixed trigger patterns are easier to detect~\cite{wang2019neural,gao2019strip} and in most cases, human visible.
    Thus, input-dependent triggers are natural inheriting from the human imperceptible requirement.
    We want to have an image perturbation function \(R(\bm{x}) = T(\bm{x}) - \bm{x}\) that satisfies
    \begin{equation}\label{eq:goal_dynamic}
        \begin{cases}
        \mathcal{M}_\theta(\bm{x} + R(\bm{x})) = \eta(y) \\
        \mathcal{M}_\theta(\bm{x} + R(\bm{x}^{\prime})) = y &
        \end{cases}\quad \text{where } \bm{x}^{\prime} \neq \bm{x}
    \end{equation}
    \item \textbf{No auxiliary training}: 
    Many existing works try to realize input-dependent attacks by utilizing an auxiliary model, e.g., DFST~\cite{cheng2020deep} uses CycleGAN.
    Such attacks are unstable because their effects depend on the training of the auxiliary models. 
    Moreover, it has high computation overhead.
    In contrast, we try to design an efficient Trojan attack without auxiliary models.
\end{itemize}

\noindent
\textbf{Adversary capabilities.}
Following existing attacks~\cite{nguyen2021wanet,doan2021lira}, we assume the adversary has full control of datasets, training process, and model implementation.
The adversary injects the Trojan by poisoning the dataset.

\subsection{Human Imperceptible Theory}\label{sec:biotheory}
Our idea of generating human imperceptible triggers is from the biology study that human visual systems are insensitive to color bit depth change.
Nadenau et al.~\cite{nadenau2000human} and many existing literatures~\cite{jacobs1991retinal,judd1952color,zeki1980representation,neitz1986polymorphism} supported this observation.
Image color quantization~\cite{heckbert1982color,bloomberg2008color,celebi2011improving,verevka1995local} is a process that reduces the number of distinct colors used in an image with the intention to produce human imperceptible changes.
To remove the unnaturalness introduced by color bit change, dithering~\cite{Floyd:1976:AAS,hu2016simple,ulichney1993void} can improve its quality.

\subsection{\bf\sys}\label{sec:squeezing}
To achieve the aforementioned objectives, we design a novel image color quantization based Trojan attack.
Spectrally, we leverage image color quantization and dithering to generate high-quality attack triggers and poisoning samples and then propose a contrastive learning and adversarial training-based method to inject the Trojan.

\noindent
\textbf{Image quantization.}
The first step of \sys is to perform image quantization, which contains two steps.
First, we squeeze the original color palette (\(m\) bits for each pixel on each channel) of the image into a smaller color palette (\(d\) bits) by reducing the color depth.
For each pixel, we use the nearest pixel value in the squeezed \(d\)-bits space to replace the original value.
The squeezing function \(T\) is defined in \autoref{eq:squeezing}, where \(round\) represents the integer rounding function:
\begin{equation}\label{eq:squeezing}
	T(\bm{x}) = \frac{round\left(\frac{\bm{x}}{2^{m}-1} *
		\left(2^{d}-1\right)\right)}{2^{d}-1}*\left(2^{m}-1\right)
\end{equation}
This is the main algorithm to generate Trojan triggers and has a few benefits.
First, it is a simple and deterministic function with good stability and generalizability, and we do not need to train any auxiliary models such as auto-encoders and U-Nets.
Second, as pointed out by existing work~\cite{xu2017feature, nadenau2000human}, large color depths are not necessary for representing images, which means the squeezed image can have high visual similarity to the original image.
While being human imperceptible, such digital value changes can be captured by ML models and used as a trigger.

\noindent
\textbf{Dithering.}
Image quantization potentially can cause unnatural regions, especially when the bit reduction is high.
To increase the stealthiness of \sys, we utilize image dithering techniques to remove the noticeable artifacts by leveraging the existing colors of the artifacts.
Image dithering techniques are designed to create the illusion of color depth when color palette of image is limited.
Specifically, we use Floyd–Steinberg dithering~\cite{Floyd:1976:AAS} and nearest-value color quantization combined with dithering as Trojan transformation.
Details are presented in \autoref{alg:dithering_0}. 
Function \(quantize\) implements \autoref{eq:squeezing}. 
Floyd–Steinberg dithering achieves its goal by error diffusion, and line 4 calculates the error.
After that, it adds residual quantization errors of a pixel onto its neighbors and spreads the debt out based on a predefined distribution.
Lines 5 to 9 implement this idea. 

\begin{algorithm}[]
    %\algsetup{linenosize=\tiny}
    % \scriptsize
 	\caption{Quantization with Floyd-Steinberg Dithering}\label{alg:dithering_0}
    {\bf Input:} % Input
    \hspace*{0.05in} Image \(I\), Diffusion Distribution \([a_1,a_2,a_3,a_4]\)\\
    {\bf Output:} % Output
    \hspace*{0.05in} Quantized Image
	\begin{algorithmic}[1]
	     \Function {Process}{$I$}

         \For{ x {\rm from right to left}}
            \For{ y {\rm from top to bottom}}
                \State \(error = quantize(I[x][y]) -I[x][y]  \)
                \State \(I[x][y] = I[x][y] + error \)
                \State \(I[x+1][y] = I[x][y] + error*a_1  \)
                \State \(I[x+1][y+1] = I[x][y] + error*a_2  \)
                \State \(I[x][y+1] = I[x][y] + error*a_3  \)
                \State \(I[x-1][y+1] = I[x][y] + error*a_4  \)
            \EndFor
         \EndFor
         \EndFunction
	\end{algorithmic}
\end{algorithm}

\noindent
\textbf{Contrastive Adversarial Training.}
As shown in \autoref{fig:trigger_compare}, image quantization based attack triggers is very close to original images.
On the one hand, this makes it hard to detect.
On the other hand, it makes training more difficult, mainly because of the small perturbations.
Existing poisoning techniques tend to use the original cross-entropy (CE) loss to train the Trojan model on benign and poisoning samples.
Due to the tiny perturbation introduced by image quantization, it is hard to converge when using the CE loss.
Moreover, existing training procedure leads to inaccurate and imprecise triggers.
As a result, reverse engineering can identify if a model has a Trojan by finding part of the trigger.
As a consequence, they are not robust attacks.
To overcome this challenge, we leverage contrastive supervised learning and adversarial training.

The whole training framework follows the contrastive learning framework, and we leverage the same loss function as described in existing work~\cite{khosla2020supervised}.
The key difference of our attack from existing contrastive learning is that in addition to existing negative sample generation methods, we also leverage adversarial example generation methods.
Specifically, we use the PGD attack to generate adversarial examples which flip the label of input from its original one to the target label to simulate the effects of our attack.
Then, we leverage them in training as negative examples.
Intuitively, this means we exclude such perturbations features as important features for the model to learn so that it can focus on the injected trigger that is image quantization and dithering described before. 
Note that the PGD attack is an optimization-based method and does not require training auxiliary models.

%% file: contents/evaluation.tex
\section{Experiments and Results}\label{sec:eval}

In this section, we evaluate \sys from different perspectives.
We first present the experiment setup, including datasets and other settings in \autoref{sec:eval_setup}.
In \autoref{sec:eval_effectiveness}, we show the effectiveness.
Then, we investigate the stealthiness of \sys by performing a human Inspection test (\autoref{sec:eval_stealthiness}). 
Furthermore, we evaluate \sys's resistance to existing defenses in \autoref{sec:eval_resistance}. 
We also conduct an ablation study of \sys in \autoref{sec:ablation}. 
In all experiments, the default bit depth is \(d=5\).

\subsection{Experiment Setup}\label{sec:eval_setup}

\noindent
\textbf{Datasets.}
We evaluate \sys on four datasets: MNIST, CIFAR-10, GTSRB and CelebA.
These datasets are regularly used in backdoor-related researches~\cite{gu2017badnets, liu2017trojaning, wang2019neural, liu2019abs, liu2018fine,gao2019strip,doan2020februus,nguyen2021wanet}. 
Details of these datasets are in~\autoref{tab:datasets}.
MNIST~\cite{lecun1998gradient} is used for hand-written digits recognition.
GTSRB~\cite{stallkamp2012man} is built for classifying different traffic signs.
CIFAR-10~\cite{krizhevsky2009learning} is a classification benchmark. 
CelebA~\cite{liu2015faceattributes} is a large-scale face attributes classification dataset. 
Note that CelebA has 40 independent binary attributes, where most attributes are unbalanced. 
To make it suitable for multi-class classification, following WaNet~\cite{nguyen2021wanet}, we use the top three most balanced attributes (i.e., Heavy
Makeup, Mouth Slightly Open, and Smiling) and concatenate them to build 8 classification classes.

\input{tf/datasets.tex}
\input{tf/effectiveness.tex}

\noindent
\textbf{Evaluation Metrics.}
Following existing works~\cite{gu2017badnets,nguyen2021wanet,li2021invisible,doan2021lira}, we use benign accuracy (BA) and attack success rate (ASR)~\cite{veldanda2020nnoculation} to evaluate the effectiveness of different Trojan attacks.
In detail, BA evaluates the accuracy of a model for clean samples by measuring the number of correctly classified clean samples over the number of all clean samples. 
ASR is the success rate of Trojan attacks.
It is defined as the number of Trojan samples that successfully perform Trojan attacks over the total number of Trojan samples.

\noindent
\textbf{Models.}
We evaluated \sys on seven popular models.
These models are commonly used in Trojan-related studies~\cite{nguyen2021wanet,doan2021lira,liu2020reflection,liu2019abs,TrojAI:online,liu2017trojaning}. First, we follow the settings of WaNet~\cite{nguyen2021wanet} and use a 5-Layer CNN (details can be found in \S{} 7.3 in Supp.) for MNIST. 
For CIFAR10 and GTSRB, we use Pre-activation ResNet18~\cite{he2016identity}. For CelebA, we use ResNet18. 
We also evaluates the effectiveness of \sys on more representative models (i.e., MobileNetV2~\cite{sandler2018mobilenetv2}, SENet18~\cite{hu2018squeeze}, ResNeXt29~\cite{xie2017aggregated} and DenseNet121~\cite{huang2017densely}).

\noindent
\textbf{Baseline.}
We select the state-of-the-art backdoor attack method WaNet~\cite{nguyen2021wanet} as baseline methods and compare the effectiveness and stealthiness with it.
The stealthiness of WaNet is much better than previous Trojan attacks~\cite{gu2017badnets,liu2017trojaning,liu2020reflection,barni2019new,chen2017targeted}, while its attack success rate is still high. 
For WaNet, We use the default hyperparameters in the original paper to conduct the attack. 
We also compare \sys with auxiliary model based method~\cite{li2021invisible} in \S{} 7.5 (Supp.).
\input{tf/different_models.tex}

\subsection{Effectiveness}\label{sec:eval_effectiveness}
To measure the effectiveness of \sys, we collect BA and ASR of \sys, benign models, and state-of-the-art baseline WaNet~\cite{nguyen2021wanet} under different datasets.
For attack settings, both all-to-one and all-to-all attacks are included.
We also evaluate \sys's generalizability to different models.
The results for all-to-one attack and all-to-all attack are shown in \autoref{tab:effectiveness_all2one} and \autoref{tab:effectiveness_all2all}, respectively.
For the all-to-one attack setting, \sys achieves higher BA and ASR than WaNet, indicating it has better performance.
In all-to-all attack settings, similarly, \sys still performs better than WaNet.
For example, the ASR of \sys is higher than that of WaNet by 0.96\%, while the BA of \sys is also higher.
These results indicate \sys is a more effective attack method.

Besides the default models used in \autoref{tab:effectiveness_all2one} and \autoref{tab:effectiveness_all2all} (i.e., a 5-Layer CNN for MNIST, Pre-activation ResNet18~\cite{he2016identity} for CIFAR-10 and GTSRB, ResNet-18 for CelebA).
We also conduct experiments on more models to further evaluate the generalizability of \sys on different network architectures (MobileNetV2~\cite{sandler2018mobilenetv2}, SENet18~\cite{hu2018squeeze}, ResNeXt29~\cite{xie2017aggregated} and DenseNet121~\cite{huang2017densely}).
The results are shown in \autoref{tab:different_models}.
In detail, we use the other four networks on CIFAR-10 and collect the ASR and BA of our method.
We also record the BA of benign models.
The attack setting is an all-to-one attack.
In all cases, \sys achieves similar BA with nearly 100\% ASR, demonstrating \sys's generalizability on different network architectures.

\input{tf/human.tex}

\input{figtex/strip_cifar10.tex}
\input{figtex/gradcam.tex}

\subsection{Stealthiness}\label{sec:eval_stealthiness}
To examine the stealthiness of different Trojan attacks, we conduct a similar human inspection study as performed in previous works~\cite{nguyen2021wanet,doan2021lira}.
We use the same settings as WaNet. 
First, 25 images are randomly selected from GTSRB~\cite{stallkamp2012man} dataset. 
Then, their corresponding Trojan images for different Trojan attack methods are created. 
For each attack method, we can get a set of 50 images by mixing the Trojan samples and original samples. 
Finally, 40 humans classify whether each image is a Trojan sample. 
Before the classifying process, the participants are trained about the attacks' characteristics and mechanisms.
The results are demonstrated in \autoref{tab:human}.
As shown in the results, \sys achieves about 50\% success fooling rate for both Trojan inputs and clean inputs, showing it has satisfying stealthiness.
WaNet~\cite{nguyen2021wanet} has higher success fooling rates than prior works.
However, as shown in \autoref{fig:trigger_compare}, it still leaves some subtle artifacts, which can be found by human insepctions.
More examples for comparing \sys and WaNet can be found in \S{} 7.1 in Supp.

\subsection{Resistance to Existing Defenses}\label{sec:eval_resistance}

To examine \sys's robustness against existing Trojan defenses, we implement representative Trojan defense methods (i.e., STRIP~\cite{gao2019strip}, GradCAM~\cite{selvaraju2017grad}, Neural Cleanse~\cite{wang2019neural} and Fine-pruning~\cite{liu2018fine}) and evaluate the resistance of \sys against them.
We also show \sys's robustness against Spectral Signature~\cite{tran2018spectral}, Universal Litmus Patterns~\cite{kolouri2020universal}, and Neural Attention Distillation~\cite{li2021neural} in \S{} 7.4 in Supplementary Materials.

\noindent
{\bf STRIP~\cite{gao2019strip}.}
We first evaluate if \sys can bypass a representative runtime Trojan attack detection method STRIP~\cite{gao2019strip}.
For a given input sample, STRIP examines if it is a Trojan sample by intentionally perturbing it via superimposing various image patterns and observing the consistency of predicted classes for perturbed inputs.
If the entropy is low (i.e., the predictions on perturbed inputs are consistent), then STRIP regard it as a Trojan sample.
\autoref{fig:strip} demonstrates the experiment results on STRIP.
The results show that the entropy range of clean models and Trojan models generated by our method are similar, indicating our attack is resistant to runtime defense STRIP.
The reason why \sys can bypass STRIP is that the superimposing operation of STRIP will modify the color distribution and break the color-shifting Trojan patterns.

\input{figtex/neural_cleanse.tex}
\input{figtex/fp.tex}

\noindent
{\bf GradCAM~\cite{selvaraju2017grad}.}
We then evaluate the robustness of \sys against GradCAM based defense methods~\cite{chou2018sentinet,doan2020februus}.
These defense mechanisms exploit GradCAM to analyze the decision process of the models.
In detail, given a model and an input sample, GradCAM can give a heatmap, where the heat value of each pixel indicates this pixel's importance for the final prediction of the model.
GradCAM is useful for detecting small-sized Trojans~\cite{gu2017badnets,liu2017trojaning}.
This is because such Trojans will produce high heat values on small-sized trigger regions, which induces abnormal GradCAM heatmap.
However, our Trojan transformation function modifies the entire image, making GradCAM fail to detect it.
\autoref{fig:gradcam} shows the visualization heatmaps of a clean model and a Trojan model generated by our method.
It shows that the heatmaps of these two models are similar, indicating \sys is resistant to GradCAM based defense methods.

\noindent
{\bf Neural Cleanse~\cite{wang2019neural}.}
We then evaluate \sys's resistance to a representative reverse engineering based defense, Neural Cleanse (NC).
It first reconstructs a trigger pattern for each class label via an optimization process.
Then, it examines if there exists a class that has significantly smaller reverse-engineered trigger and considers it as a sign of Trojan models.
In detail, it uses Anomaly Index (i.e., Median Absolute Deviation~\cite{hampel1974influence}) to quantify the deviation of reverse-engineered triggers based on their sizes and consider the models whose Anomaly Index is larger than two as Trojan models.
Although it is effective for detecting patched-based Trojans~\cite{gu2017badnets,liu2017trojaning}, it assumes that different samples share the same trigger pattern in pixel level.
Our method can bypass NC by breaking this assumption with \emph{Input-dependent} triggers, i.e., the pixel level Trojan perturbations for different samples are different.
Experiment results shown in \autoref{fig:neural_cleanse} demonstrate Neural Cleanse fails to detect the Trojan model generated by our method.

\noindent
{\bf Fine-pruning~\cite{liu2018fine}.}
We then investigate \sys's resistance to representative Trojan removing method, Fine-pruning.
This defense is based on the assumption that Trojan behaviors are related to a few dormant neurons in the model, and the Trojan can be removed via pruning such dormant neurons.
Given a set of clean samples, it records the activation values on a layer and considers the neuron that has the smallest activation value as the most dormant neurons.
Then, it gradually prunes neurons based on the order of their activation values.
The results can be found in \autoref{fig:fp}.
It shows that Fine-pruning is not able to remove the Trojan injected by our methods.
For example, in MNIST, CIFAR-10, and GTSRB, the ASR is always close or higher than BA.
For CelebA, although the ASR is slightly lower but it still achieves above 50\%, meaning the Trojan is not completely removed.

\subsection{Ablation Study}\label{sec:ablation}
To investigate the effects of hyperparameters and different components, we first evaluate the effects of the bits number \(d\).
Then, we study the influence of different injection rates.
We also investigate the effects of dithering and contrastive adversarial training.

\noindent
{\bf Bits Number.}
As mentioned in \autoref{sec:squeezing}, to generate Trojan samples, we quantize the original color palette (\(m\) bits for a pixel on each channel) into a smaller color palette (\(d\) bits), and use the nearest pixel value in the squeezed value space to replace the original one.
Here, the bits number of the squeezed color palette \(d\) is called bits number.
To investigate the effects of different bits number \(d\), we collect the BA and ASR under different bits numbers.
The used dataset is CIFAR-10, and the attack setting is an all-to-one attack.
We also show the generated Trojan sample to study bits number's influence on the stealthiness of the attack.
\autoref{fig:depth} shows the BA and ASR under different bits
number \(d\).
The results demonstrate that our method can achieve high BA and high ASR when \(d\) is not larger than 6.
However, when \(d\) reaches 7, the ASR decreases.
Note that the original images' bits number for each pixel on each channel is 8. The larger \(d\) is, the fewer perturbations the attack induces.
When \(d = 7\), the difference between the Trojan sample and the benign sample is so small that it is hard for the model to tell.
\autoref{fig:different_squeeze} demonstrates the generated Trojan samples under
different bits number \(d\).
For different \(d\) values, the Trojan sample is natural and indistinguishable from the clean sample.
More examples generated under different bits number \(d\) can be found in \S{} 7.2 in Supp.

\noindent
{\bf Injection Rate.}
During training, the model is optimized on benign samples and Trojan samples alternatively.
We denote the fraction that the model is optimized on Trojan samples as injection rate \(\alpha\).
To investigate its influence on \sys's performance, we record the BA and ASR with different injection rates.
The used dataset is CIFAR-10, and the attack setting is an all-to-one attack.
The results are shown in \autoref{fig:poisoning_rate}.
The ASRs are low when \(\alpha\) is small.
This is because a small injection rate indicates the effects of optimizing on Trojan sample and target labels is limited so that the model fails to learn the Trojan behaviors.
With the increase of the \(\alpha\), the ASR becomes higher.
BA is not influenced by injection rate, when injection rate is in a range from 2.5\% to 30\%.

\noindent
{\bf Dithering.}
As we mentioned in \autoref{sec:squeezing}, when \(d\) is small, the new images can be less stealthy.
To make the Trojan samples more natural, we use dithering techniques to remove these unnatural artifacts.
Here we study the effects of dithering by illustrating the Trojan samples generated with dithering and without it.
\autoref{fig:dither} demonstrates examples to show the effects of dithering, using the GTSRB dataset as an example.
The dithering technique helps generate more natural Trojan samples by fixing the color banding.
Overall, dithering can remove the color banding artifacts in the directly quantized image to make the attack more stealthy.

\input{figtex/ablations.tex}
\input{figtex/different_depth.tex}
\input{figtex/dither.tex}

\noindent
{\bf Contrastive Adversarial Training.}
In this section, we conduct an ablation study to investigate the effects of Contrastive Adversarial Training.
We use the vanilla and our training methods to train two models on CIFAR-10, and compare them by using a trigger reverse engineering method, Neural Cleanse~\cite{wang2019neural}.
\autoref{fig:nc_ablation} shows the result. 
As we can see, the model trained with the vanilla method has an anomaly index that is higher than the threshold (i.e., 2).
By contrast, the model trained with our method successfully bypasses the detection.

\input{figtex/nc_ablation.tex}

%% file: tf/datasets.tex
\begin{table}[b]
    \centering
    \scriptsize
    \setlength\tabcolsep{4pt}
    
    \begin{tabular}{@{}ccccc@{}}
        \toprule
        Dataset     & Input Size & \#Train & \#Test  & Classes \\ \midrule
        MNIST       & 28*28*1    & 60000         & 10000        & 10      \\
        CIFAR-10    & 32*32*3    & 50000         & 10000        & 10      \\
        GTSRB       & 32*32*3    & 39209         & 12630        & 43      \\
        CelebA & 64*64*3  & 162770          & 19962         & 8      \\ \bottomrule
    \end{tabular}
    \caption{Overview of datasets.}\label{tab:datasets}
    \end{table}

%% file: tf/effectiveness.tex
\begin{table*}[]
    \begin{minipage}{0.5\linewidth}
        \centering
        \scriptsize
        \setlength\tabcolsep{4pt}
        \scalebox{1}{\begin{tabular}{@{}clcccccccl@{}}
            \toprule
            \multirow{2}{*}{Dataset} &  & Non-attack &  & \multicolumn{2}{c}{WaNet} &  & \multicolumn{2}{c}{BppAttack} &  \\ \cmidrule(lr){3-3} \cmidrule(lr){5-6} \cmidrule(lr){8-9}
                                     &  & BA         &  & BA          & ASR         &  & BA          & ASR        &  \\ \midrule
            MNIST                    &  & 99.67\%    &  & 99.52\%     & 99.86\%     &  & 99.36\%     & 99.79\%    &  \\
            CIFAR-10                 &  & 94.88\%    &  & 94.15\%     & 99.55\%     &  & 94.54\%     & 99.91\%    &  \\
            GTSRB                    &  & 99.31\%    &  & 98.97\%     & 98.78\%     &  & 99.25\%     & 99.96\%    &  \\
            CelebA                   &  & 79.14\%    &  & 78.99\%     & 99.33\%     &  & 79.06\%     & 99.99\%    &  \\ \bottomrule
            \end{tabular}
            }
            \caption{Effectiveness on all-to-one attacks.}\label{tab:effectiveness_all2one}\end{minipage}
        \begin{minipage}{0.5\linewidth}
                \centering
                \scriptsize
                \setlength\tabcolsep{4pt}

                \scalebox{1}{\begin{tabular}{@{}clcccccccl@{}}
                    \toprule
                    \multirow{2}{*}{Dataset} &  & Non-attack &  & \multicolumn{2}{c}{WaNet} &  & \multicolumn{2}{c}{BppAttack} &  \\ \cmidrule(lr){3-3} \cmidrule(lr){5-6} \cmidrule(lr){8-9}
                                             &  & BA         &  & BA          & ASR         &  & BA          & ASR        &  \\ \midrule
                    MNIST                    &  & 99.67\%    &  & 99.44\%     & 95.90\%     &  & 99.25\%     & 98.46\%    &  \\
                    CIFAR-10                 &  & 94.88\%    &  & 94.43\%     & 93.36\%     &  & 94.73\%     & 94.32\%    &  \\
                    GTSRB                    &  & 99.52\%    &  & 99.39\%     & 98.31\%     &  & 99.46\%     & 99.29\%    &  \\
                    CelebA                   &  & 79.14\%    &  & 78.73\%     & 78.58\%     &  & 78.84\%     & 78.72\%    &  \\ \bottomrule
                    \end{tabular}}
                    \caption{Effectiveness on all-to-all attacks.}\label{tab:effectiveness_all2all}\end{minipage}
    \end{table*}

%% file: tf/different_models.tex
\begin{table}[]
    \centering
    \scriptsize
    \setlength\tabcolsep{4pt}
    
    \begin{tabular}{@{}ccccccc@{}}
        \toprule
        \multirow{2}{*}{Network} &  & Non-attack &  & \multicolumn{2}{c}{BppAttack} &  \\ \cmidrule(lr){3-3} \cmidrule(lr){5-6}
                                 &  & BA         &  & BA         & ASR         &  \\ \midrule
        MobileNetV2              &  & 94.21\%    &  & 93.79\%    & 99.99\%     &  \\
        SENet18                  &  & 94.79\%    &  & 94.49\%    & 99.98\%     &  \\
        ResNeXt29                &  & 94.83\%    &  & 94.68\%    & 99.97\%     &  \\
        DenseNet121              &  & 95.35\%    &  & 95.20\%     & 100.00\%    &  \\ \bottomrule
        \end{tabular}
        \caption{Effectiveness on different networks.}\label{tab:different_models}
    \end{table}

%% file: tf/human.tex
\begin{table}[]
    \centering
    \scriptsize
    \setlength\tabcolsep{4pt}
    
    \scalebox{1}{
    \begin{tabular}{@{}ccccccc@{}}
        \toprule
        Images  & Patched & Blended & SIG   & ReFool & WaNet  & BppAttack \\ \midrule
        Trojan & 4.2\%   & 2.3\%   & 1.7\% & 5.2\%  & 42.0\% &  50.7\%         \\
        Clean  & 5.9\%   & 7.2\%  & 2.8\% & 14.5\% & 21.8\% &  48.1\%         \\
        Both    & 5.0\%   & 4.7\%   & 2.2\% & 9.8\%  & 30.9\% &   49.4\%        \\ \bottomrule
        \end{tabular}
    }
    \caption{Success fooling rates of each Trojan attacks.}\label{tab:human}
    \end{table}

%% file: figtex/strip_cifar10.tex
\begin{figure*}[]
    \centering
    \footnotesize
    \begin{subfigure}[t]{0.49\columnwidth}
        \centering
        \footnotesize
        \includegraphics[width=\columnwidth]{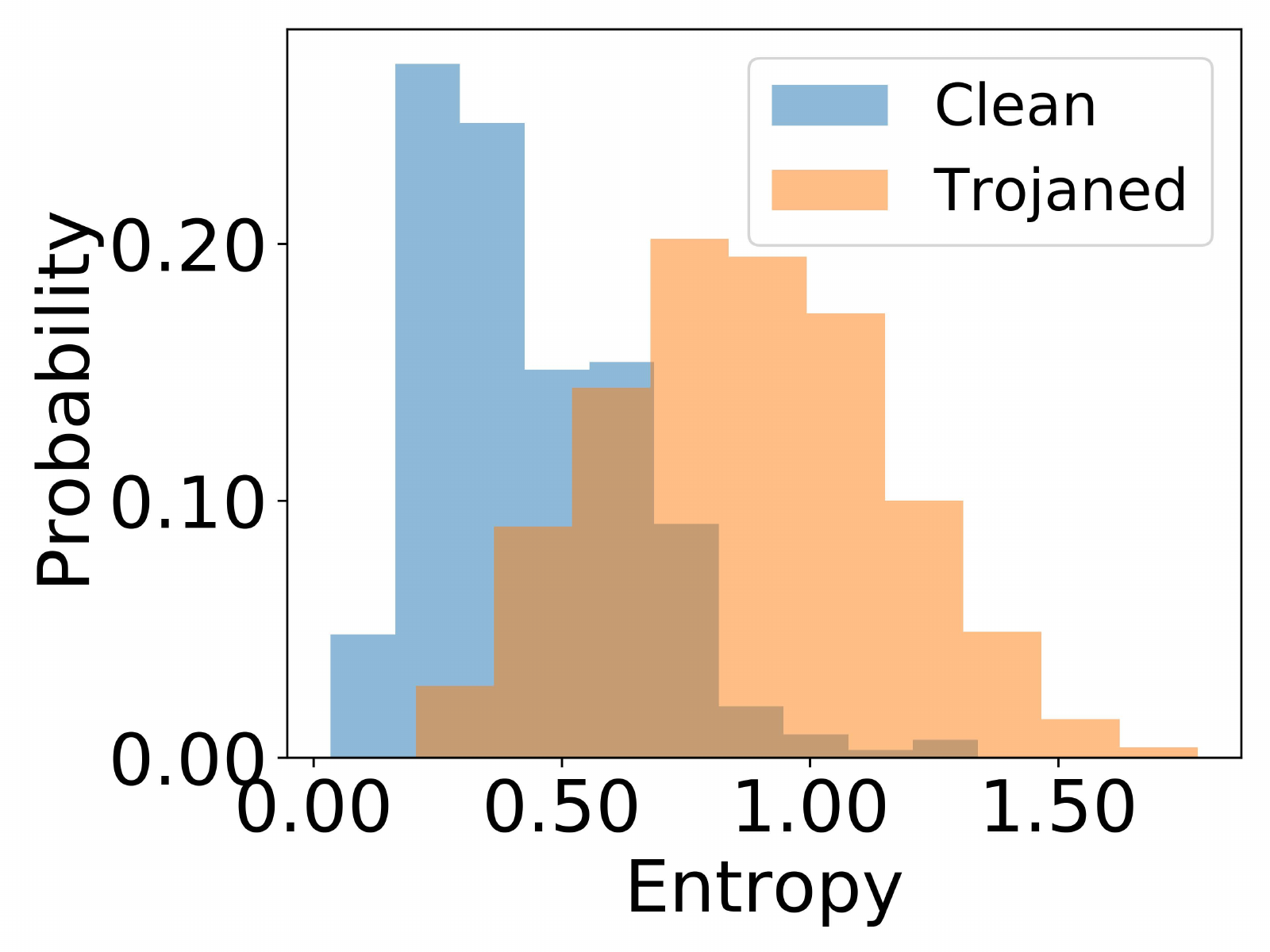}
        \caption{MNIST}        \label{fig:hidden_trigger}
    \end{subfigure}
    \begin{subfigure}[t]{0.49\columnwidth}
        \centering
        \footnotesize
        \includegraphics[width=\columnwidth]{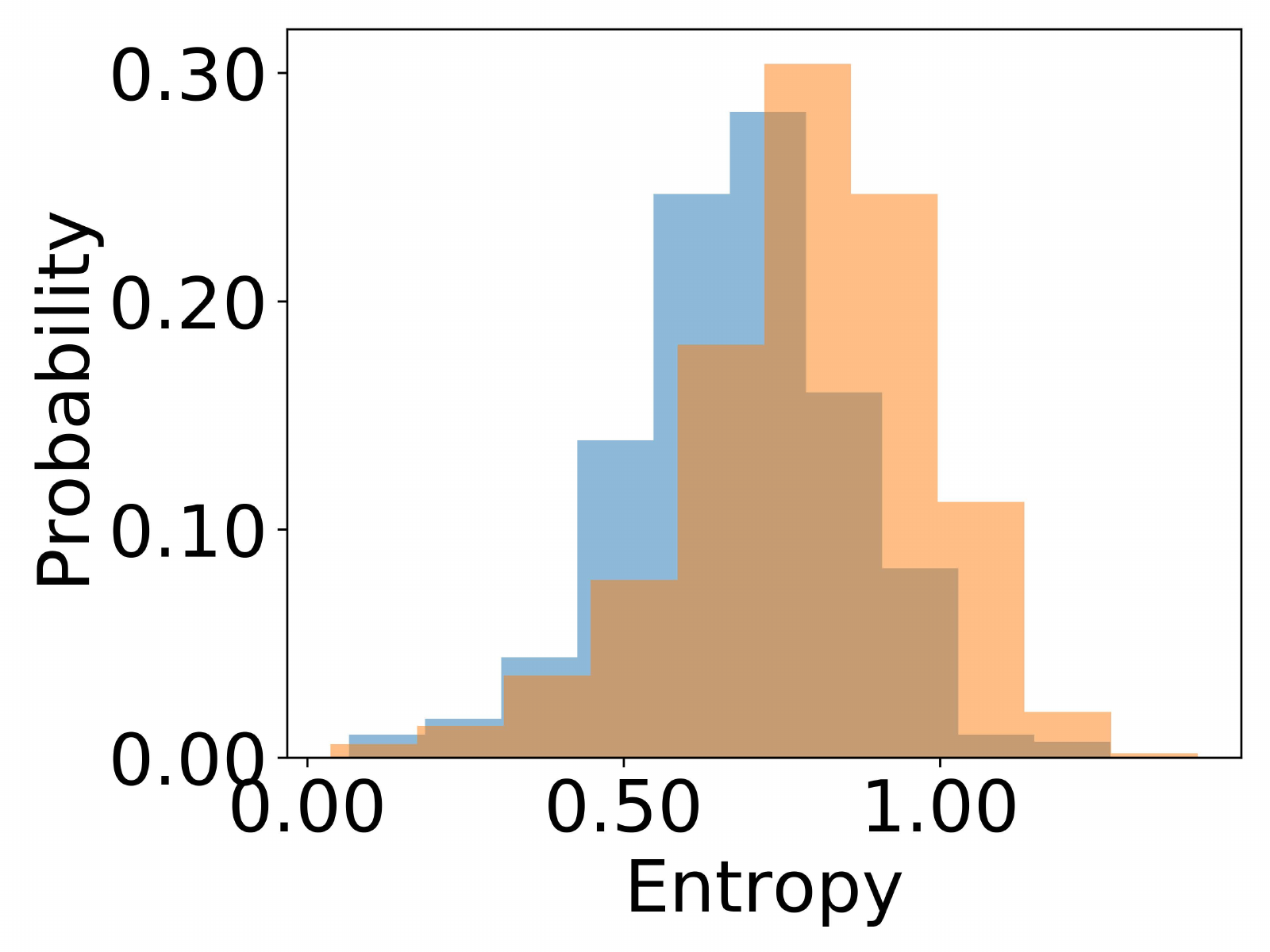}
        \caption{CIFAR-10}        \label{fig:hidden_stamp2}
    \end{subfigure}
    \begin{subfigure}[t]{0.49\columnwidth}
        \centering
           \footnotesize
           \includegraphics[width=\columnwidth]{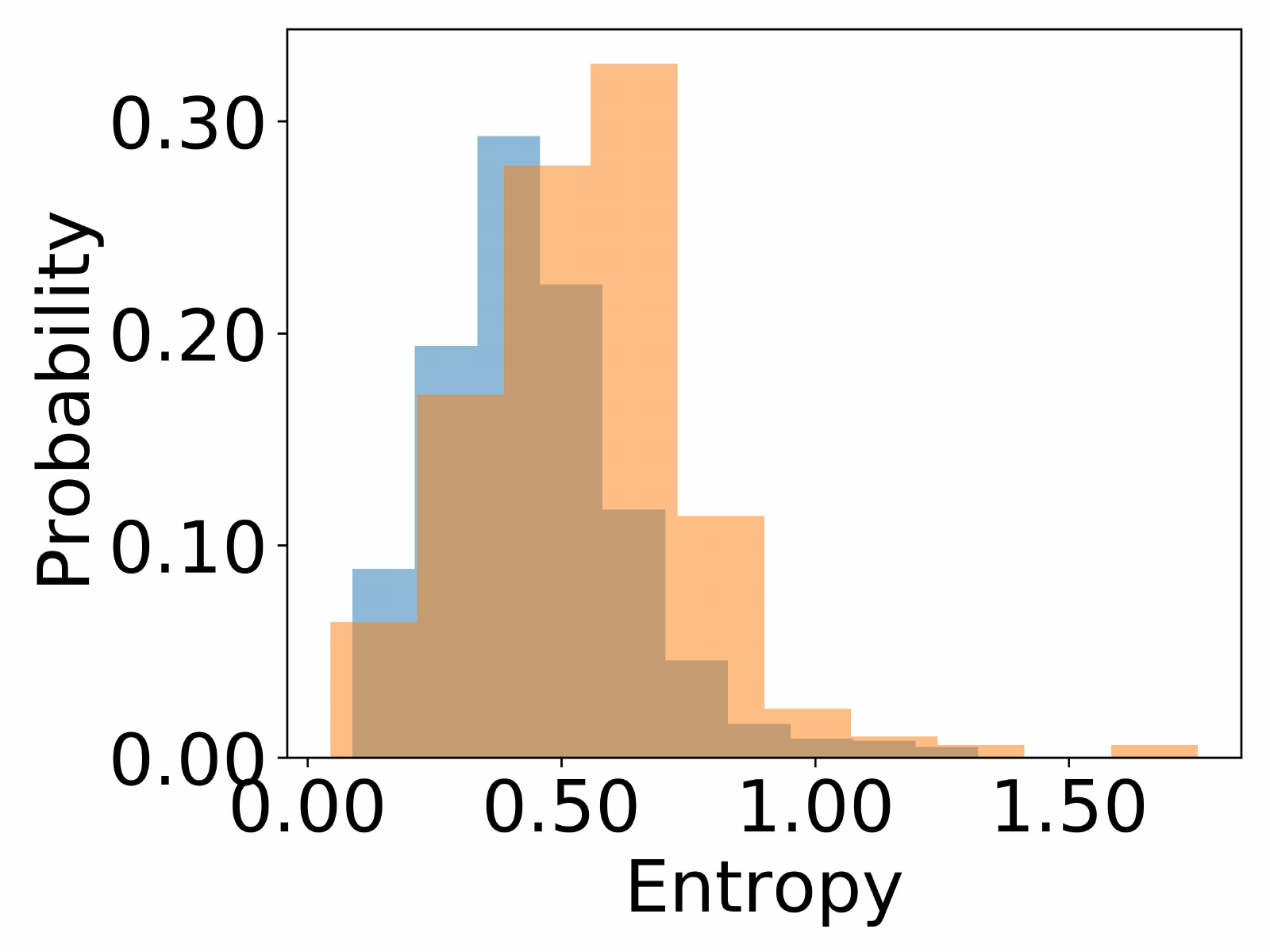}
           \caption{GTSRB}           \label{fig:reflect_natrual_trigger}
    \end{subfigure}
    \begin{subfigure}[t]{0.49\columnwidth}
        \centering
           \footnotesize
           \includegraphics[width=\columnwidth]{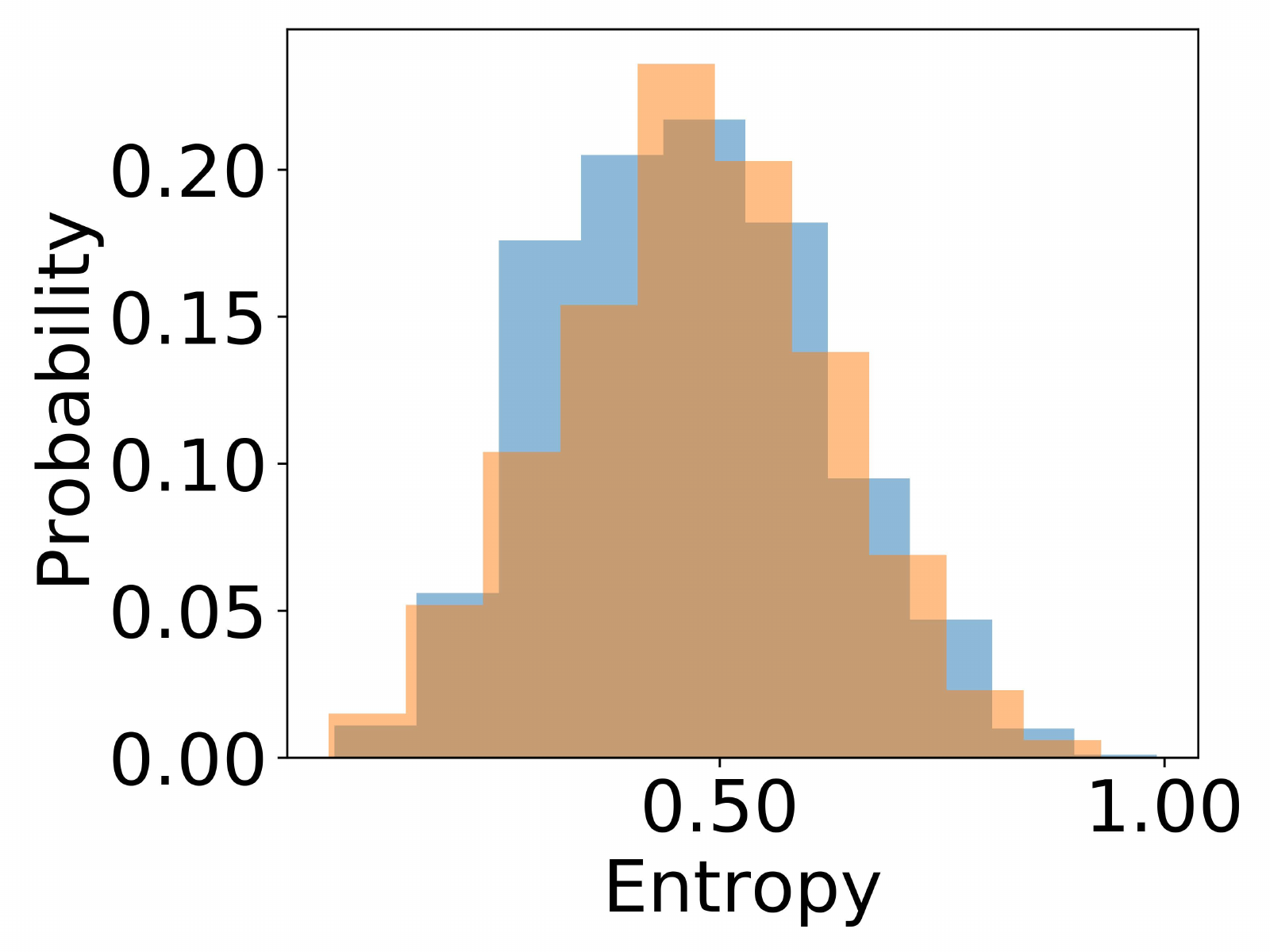}
           \caption{CelebA}           \label{fig:imagenette}
    \end{subfigure}
    \caption{Resilient to STRIP~\cite{gao2019strip}.}\label{fig:strip}
\end{figure*}

%% file: figtex/gradcam.tex
\begin{figure}[]
	\centering
	\footnotesize
	\includegraphics[width=0.7\columnwidth]{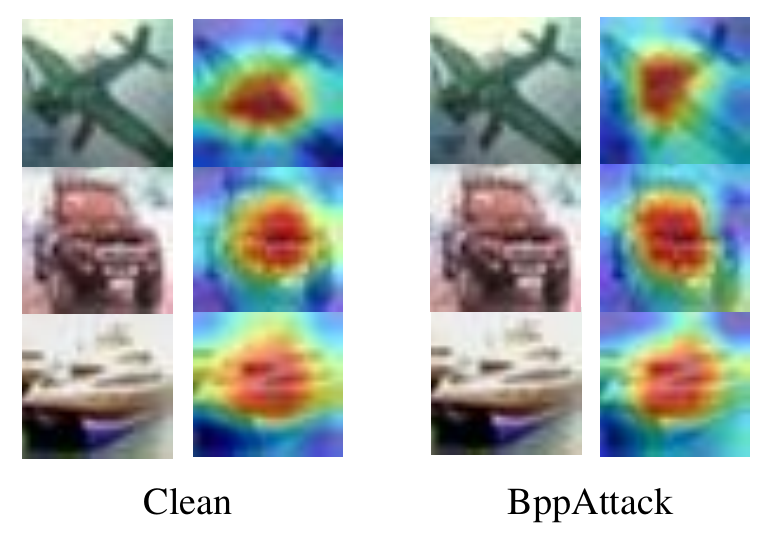}
	\caption{Resilient to GradCAM~\cite{selvaraju2017grad}.}\label{fig:gradcam}
\end{figure}

%% file: figtex/neural_cleanse.tex
\begin{figure}[]
	\centering
	\footnotesize
	\includegraphics[width=0.6\columnwidth]{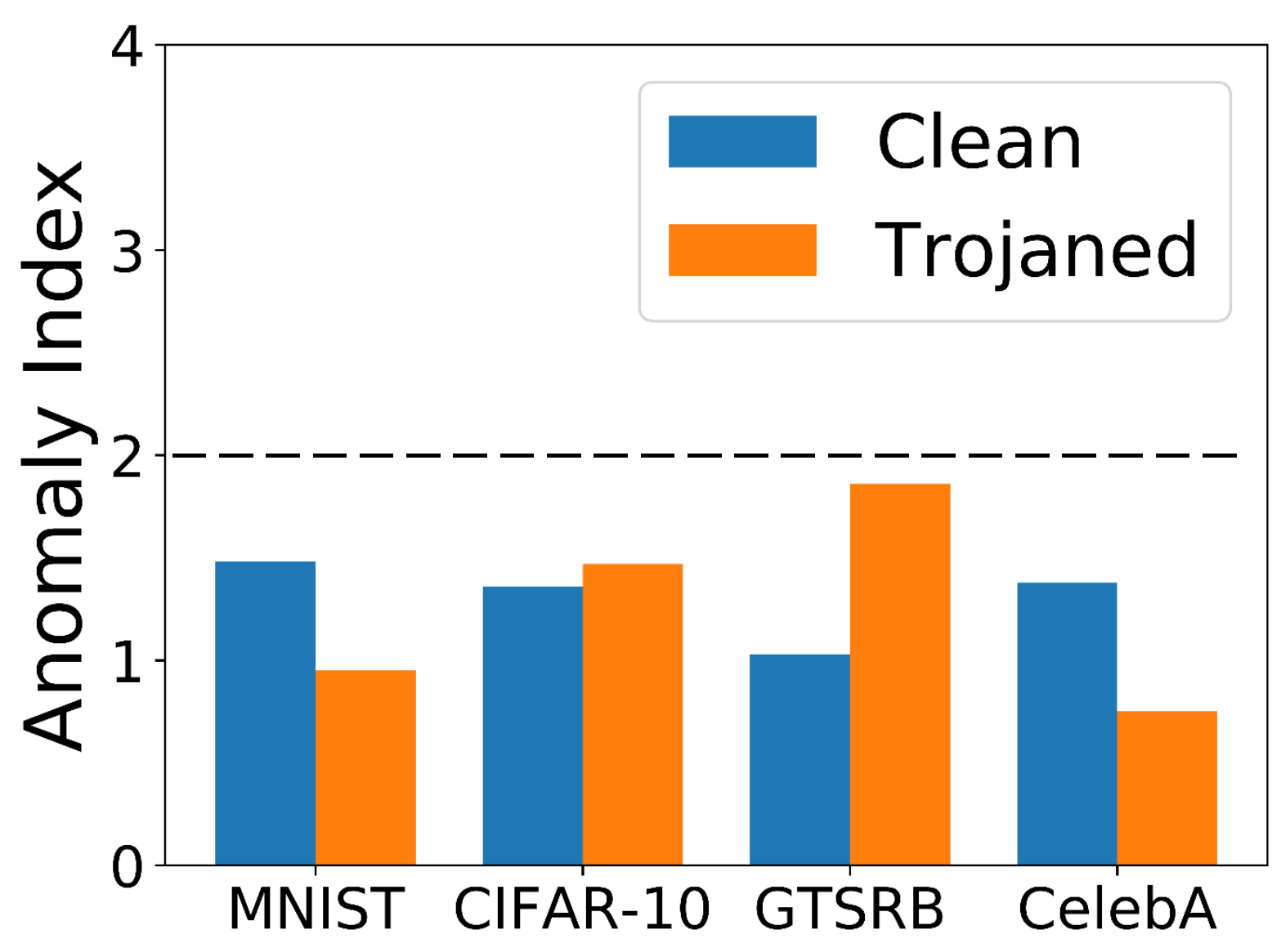}
	\caption{Resilient to Neural Cleanse~\cite{wang2019neural}.}\label{fig:neural_cleanse}
\end{figure}

%% file: figtex/fp.tex
\begin{figure*}[]
    \centering
    \footnotesize
    \begin{subfigure}[t]{0.49\columnwidth}
        \centering
        \footnotesize
        \includegraphics[width=\columnwidth]{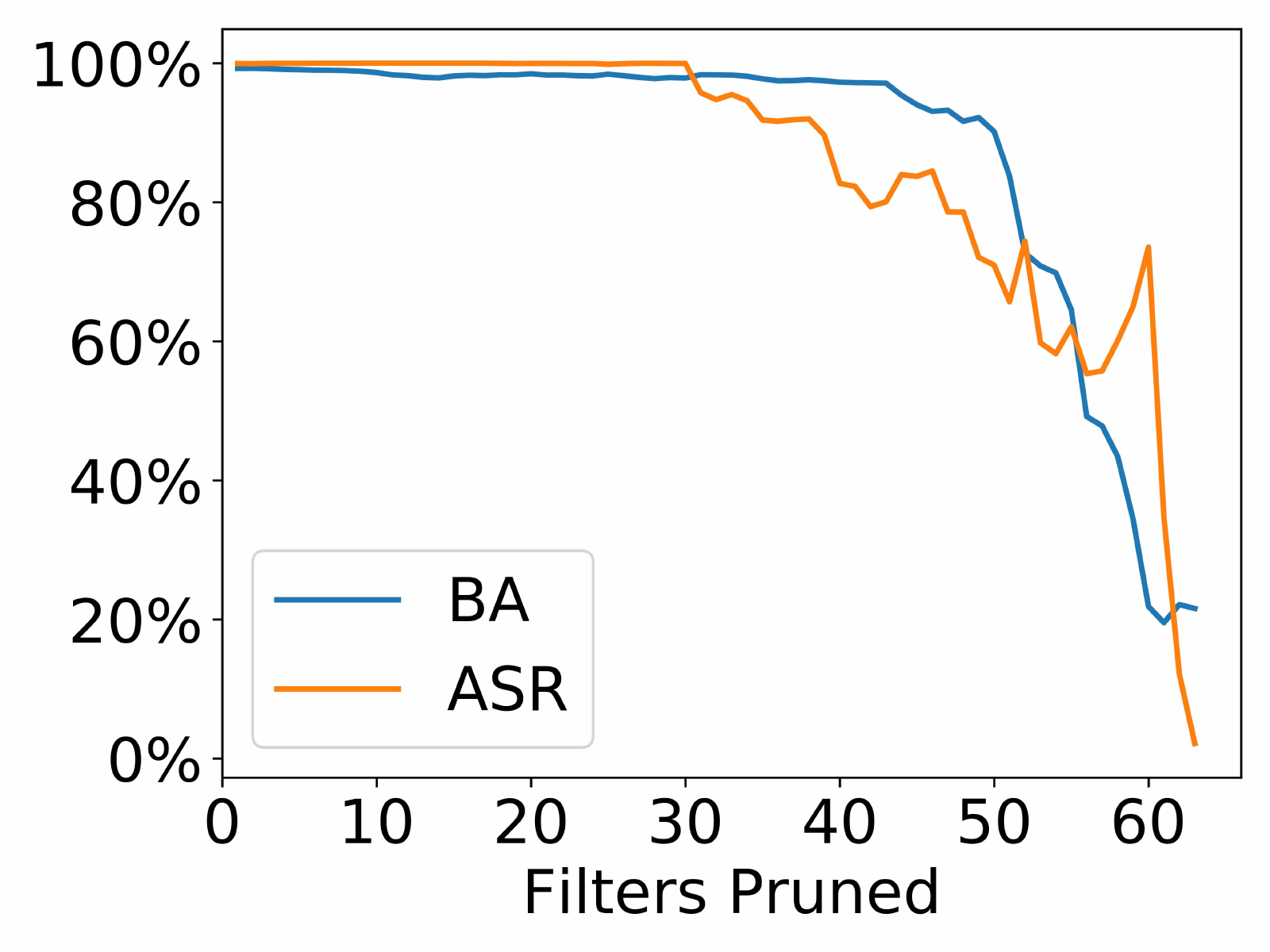}
        \caption{MNIST}
        \label{fig:hidden_trigger}
    \end{subfigure}
    \begin{subfigure}[t]{0.49\columnwidth}
        \centering
        \footnotesize
        \includegraphics[width=\columnwidth]{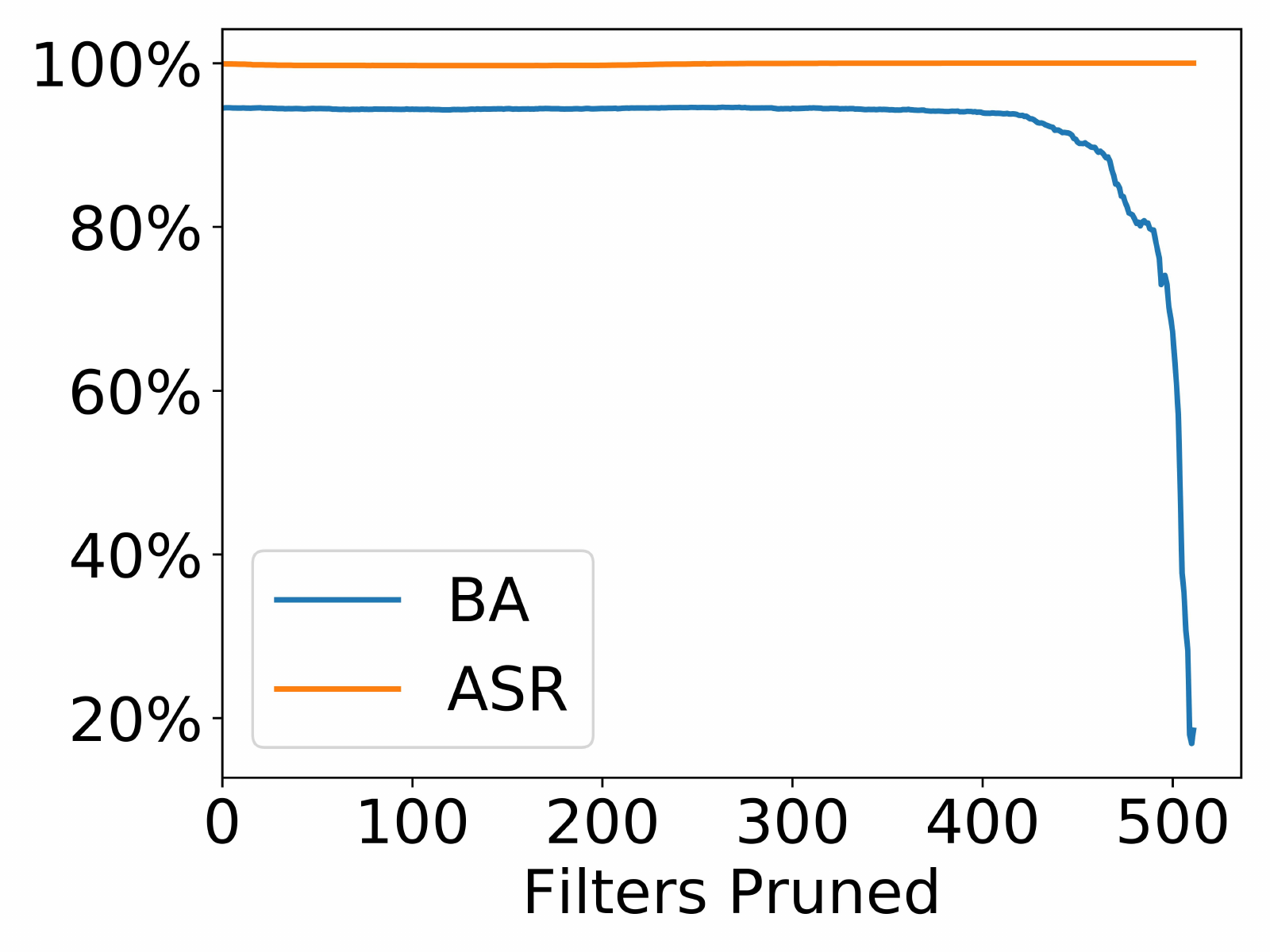}
        \caption{CIFAR-10}
        \label{fig:hidden_stamp2}
    \end{subfigure}
    \begin{subfigure}[t]{0.49\columnwidth}
        \centering
           \footnotesize
           \includegraphics[width=\columnwidth]{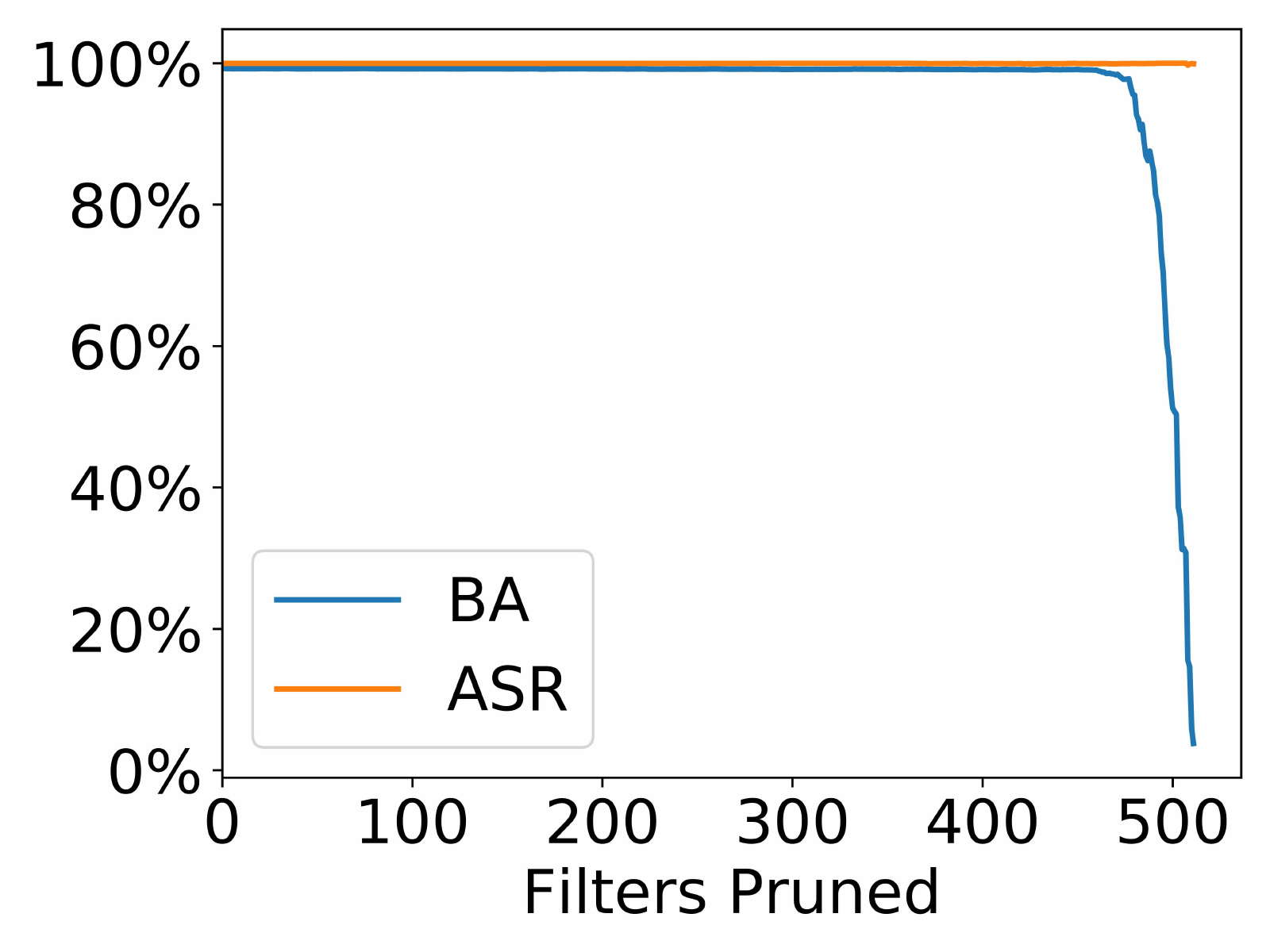}
           \caption{GTSRB}
           \label{fig:reflect_natrual_trigger}
    \end{subfigure}
    \begin{subfigure}[t]{0.49\columnwidth}
        \centering
           \footnotesize
           \includegraphics[width=\columnwidth]{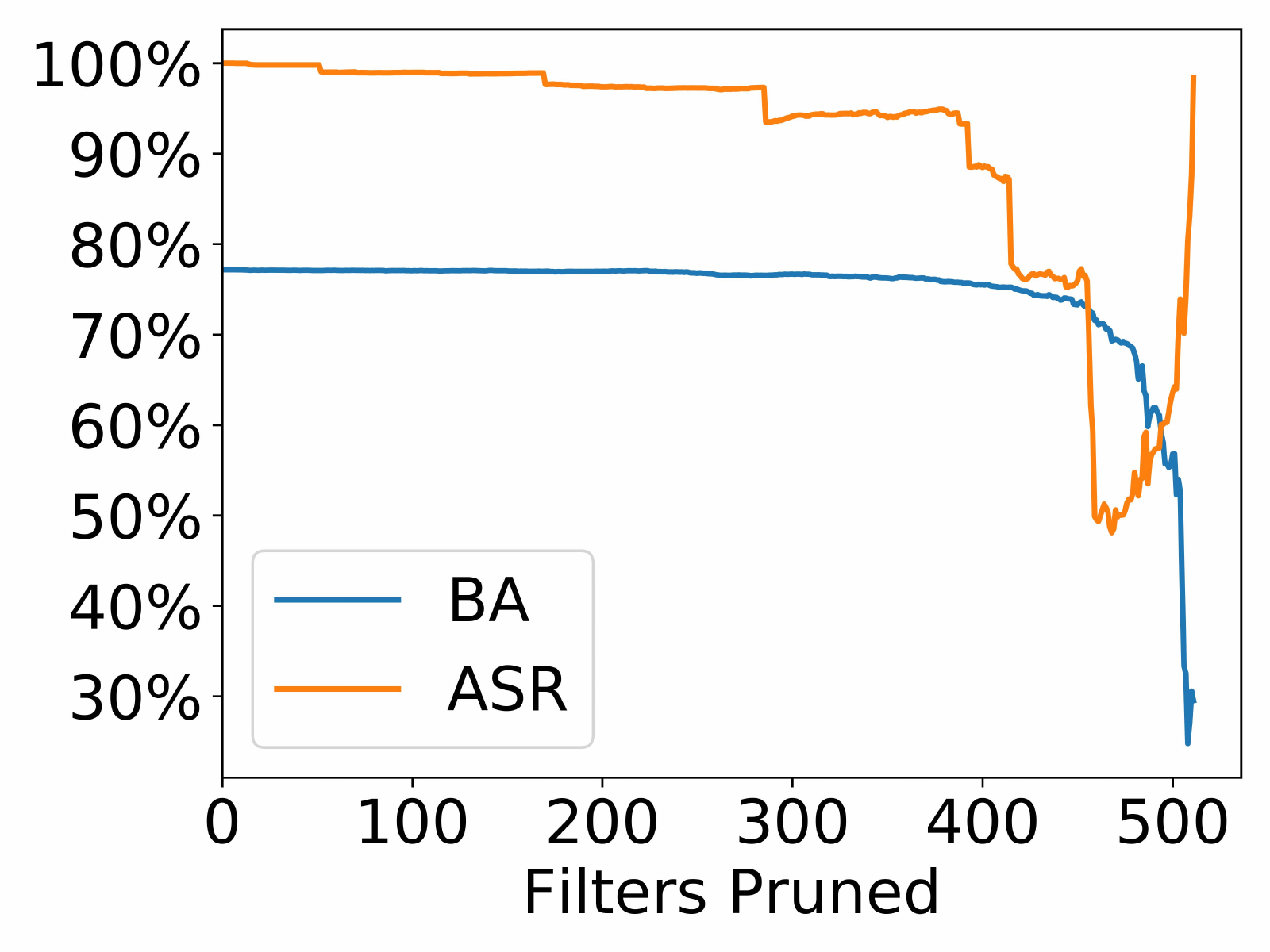}
           \caption{CelebA}
           \label{fig:imagenette}
    \end{subfigure}
    \caption{Resilient to Fine-Pruning~\cite{liu2018fine}.}\label{fig:fp}
\end{figure*}

%% file: figtex/ablations.tex
\begin{figure}[]
    \centering
    \footnotesize
    \begin{subfigure}[t]{0.49\columnwidth}
        \centering     
        \footnotesize
        \includegraphics[width=\columnwidth]{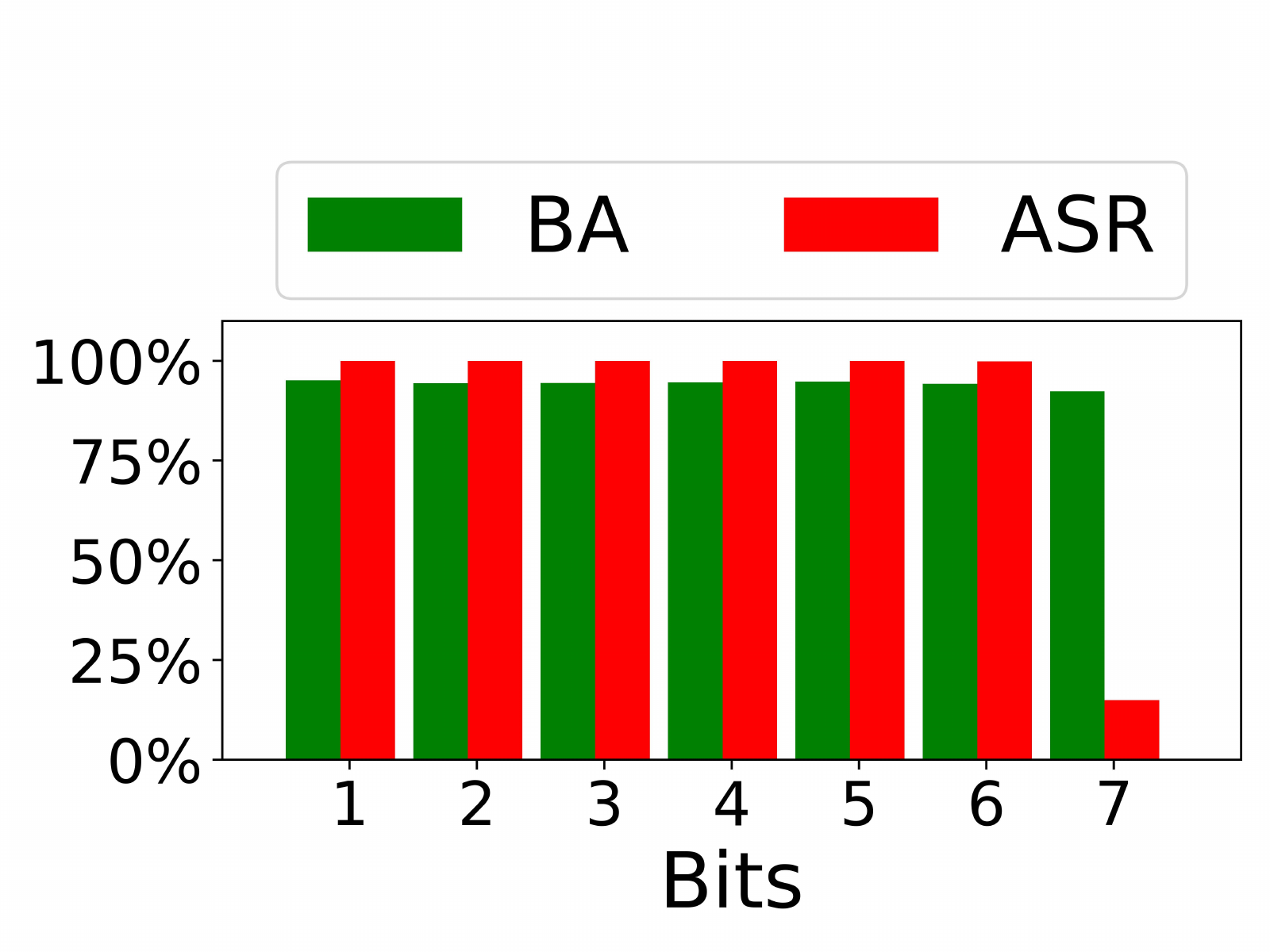}
        \caption{Performance with Different \(d\)}
        \label{fig:depth}
    \end{subfigure}
    \begin{subfigure}[t]{0.49\columnwidth}
        \centering
        \footnotesize
        \includegraphics[width=\columnwidth]{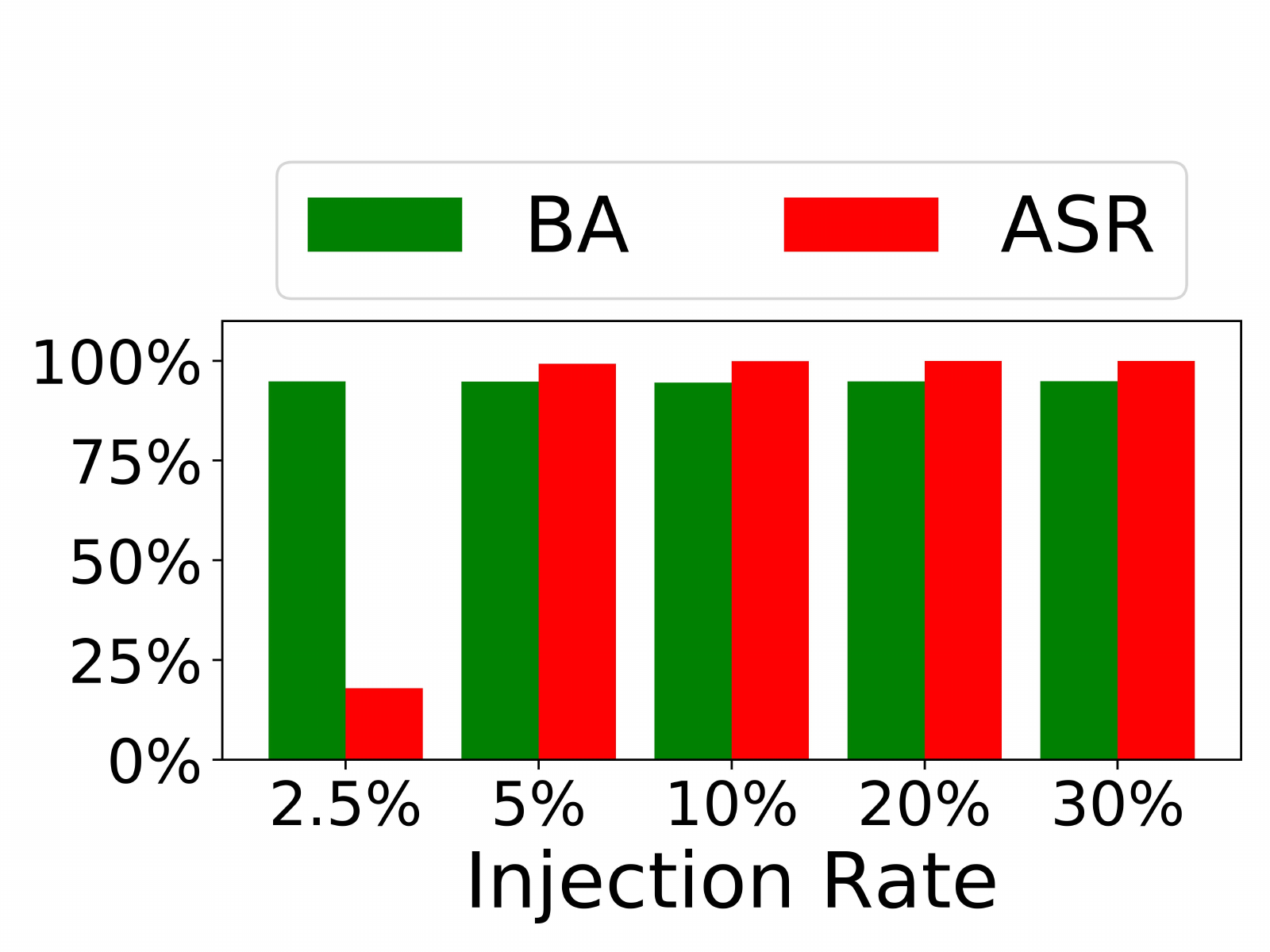}
           \caption{Performance with Different \(\alpha\)}
        \label{fig:poisoning_rate}
    \end{subfigure}
    \caption{Evaluation results with different hyperparameters.}
    \label{fig:lr}
\end{figure}

%% file: figtex/different_depth.tex
\begin{figure}[]
	\centering
	\footnotesize
	\includegraphics[width=1\columnwidth]{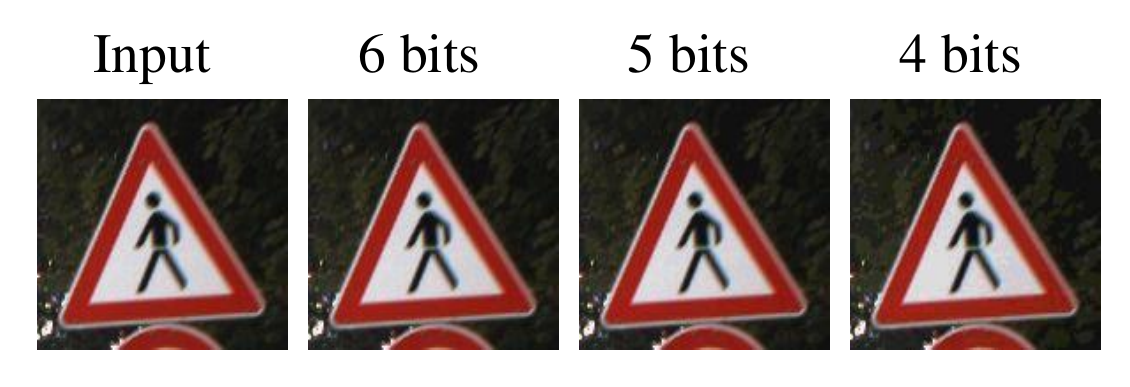}
	\caption{Effects of different bits number. }\label{fig:different_squeeze}
\end{figure}

%% file: figtex/dither.tex
\begin{figure}[]
	\centering
	\footnotesize
	\includegraphics[width=0.85\columnwidth]{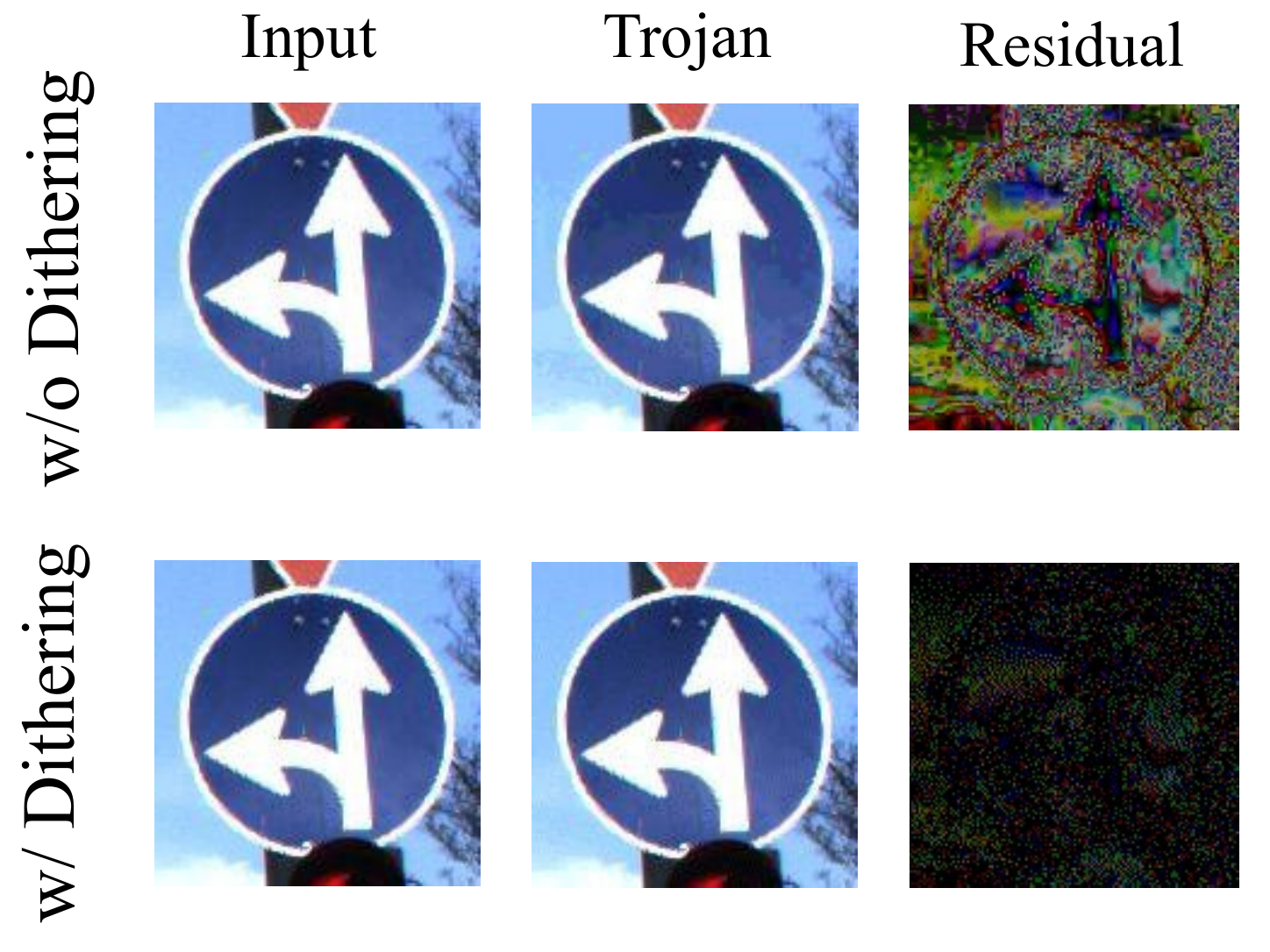}	\caption{Effects of Dithering.}\label{fig:dither}
\end{figure}

%% file: figtex/nc_ablation.tex
\begin{figure}[]
	\centering
	\footnotesize
	\includegraphics[width=0.75\columnwidth]{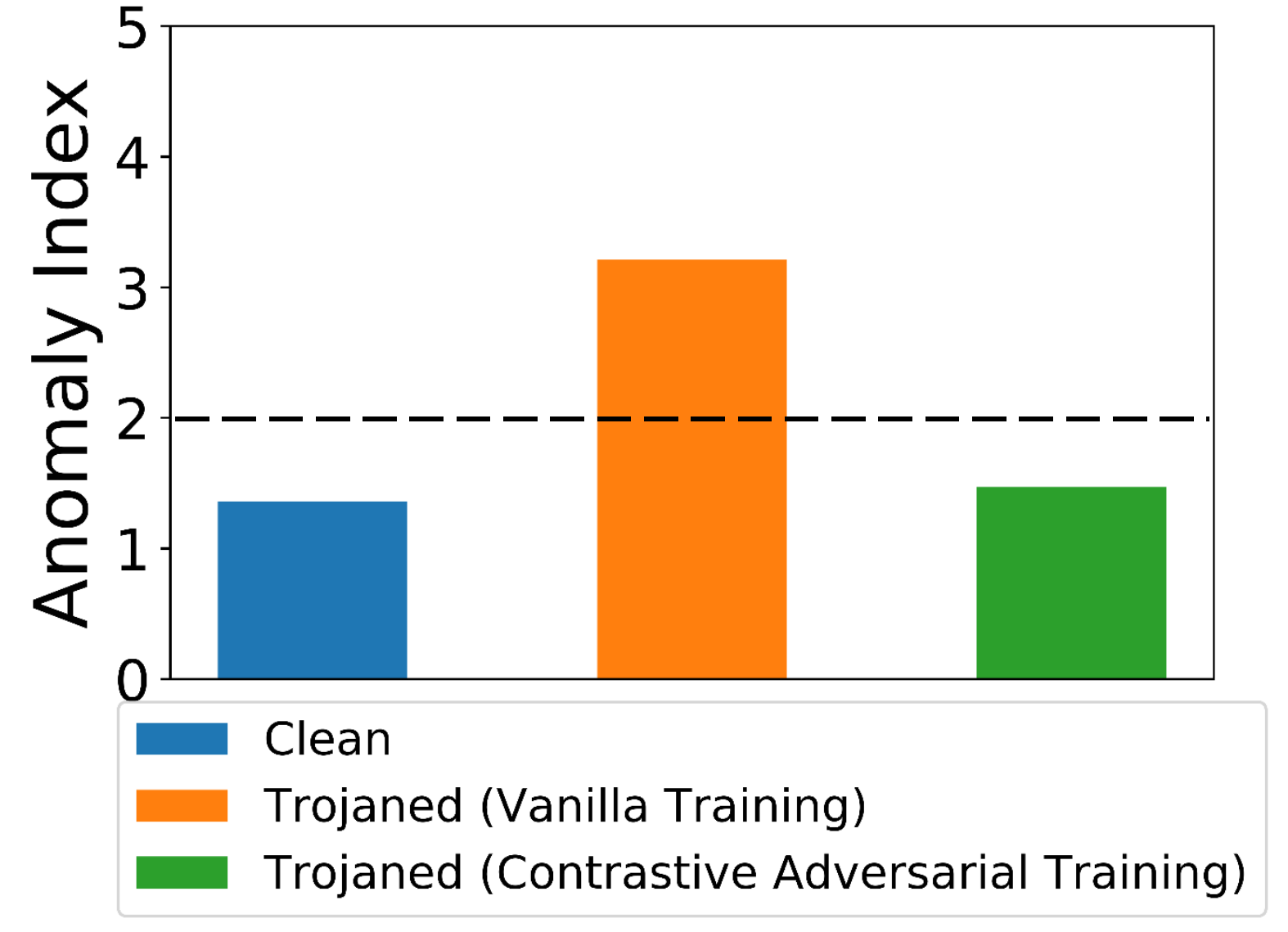}
	\caption{Effects of Contrastive Adversarial Training. The model trained by vanilla training method can be detected by Neural Cleanse, while the model trained by Contrastive Adversarial Training can bypass the detection. }\label{fig:nc_ablation}
\end{figure}

%% file: contents/discussion.tex
\section{Discussion}\label{sec:discussion}

\noindent
\textbf{Mitigations.}
\sys can bypass existing defenses, but it is not perfect.
We believe that a defense that focuses on color depth checking can potentially detect our attacks.
Other possible defenses, e.g., activation distribution checking and anomaly detection based methods can also help mitigate such attacks.
Also, it is possible to defend our attack under different threat models.
For example, data cleaning and validation or enforcing another training protocol can mitigate general data poisoning based attacks.
Recent works have proposed DP-SGD and other methods to defend such attacks~\cite{hong2020effectiveness,du2019robust} during training time.
Such methods can potentially help mitigate \sys.

\noindent
{\bf Ethical statements.}
In this paper, we propose a stealthy and efficient Trojan attack, demonstrating a threat.
On the one hand, it has potential negative societal impacts.  
The adversaries can exploit real-world AI systems, such as facial recognition applications. 
On the other hand, we disclose new vulnerabilities and alert the defenders to pay attention to such new types of Trojan attacks.

%% file: contents/conclusion.tex
\section{Conclusion}\label{sec:conclusion}

In this paper, we propose an image quantization and dithering based Trojan attack.
By exploiting the human visual system, our method can generate human imperceptible triggers with the support of literature from biology.
To improve the effectiveness of our attack, we also propose a contrastive learning and adversarial training based poisoning method.
Results show that our attack is highly effective and efficient.

%% file: contents/ack.tex
\section*{Acknowledgement}
We thank the anonymous reviewers for their constructive comments.
This work is supported by IARPA TrojAI W911NF-19-S-0012.
Any opinions, findings, and conclusions expressed in this paper are those of the authors only and do not necessarily reflect the views of any funding agencies.

%% file: contents/appendix.tex
%\appendix

\setcounter{section}{6}
\section{Supplementary materials}\label{sec:appendix}

\subsection{Additional Images for Different Attacks}
\input{figtex/appendix_diff_attacks.tex}
In this section, we show more Trojan samples generated by WaNet~\cite{nguyen2021wanet} and our \sys.
The images can be found in \autoref{fig:appendix_diff_attacks}, where the first row is the original images, and the second and the third row are the Trojan samples generated by WaNet and \sys, respectively.
As can be observed, the Trojan
samples generated by WaNet can be spotted, and \sys is more stealthy.

\subsection{Additional Images for Different Bits Numbers}
To illustrate the effects of different bit numbers, in this section, we demonstrate more samples generated by different bits numbers.
The results are shown in \autoref{fig:appendix_diff_bits}.
It shows that the Trojan samples produced by \sys with different bits numbers are natural and stealthy.

\subsection{Details of MNIST Classifier}
The detailed architecture of the classifier used for MNIST dataset is shown in \autoref{tab:mnist_classifier}.

\input{tf/mnist_classfier.tex}

\input{figtex/appendix_diff_bits.tex}

\subsection{Resistance to More Defenses}

\noindent
\textbf{Spectral Signature~\cite{tran2018spectral}.}
Spectral Signature~\cite{tran2018spectral} is a defense method that identifies and removes Trojans during training.
Although it is a training time defense and does not match our threat model, investigating if the Trojan samples generated by \sys can be detected by it is still helpful.
Given a set of benign and Trojan samples, Spectral Signature first collects the latent features and computes the top singular value of the covariance matrix.
Then, for each sample, it calculates the correlation score between its features and the top singular value that is used as the outlier scores.
Finally, it removes the samples with high outlier scores.
We use 900 benign samples and 100 Trojan samples in CIFAR-10 to evaluate if our attack can bypass Spectral Signature.
The results are demonstrated in \autoref{fig:ss}.
It shows that we can fool the detector and bypass the detection.

\input{figtex/ss.tex}

\noindent
\textbf{Universal Litmus Patterns~\cite{kolouri2020universal}.}
ULP~\cite{kolouri2020universal} is designed to detect if a model is Trojan or not.
It first trains universal patterns from a large number of benign and Trojan models. 
These patterns are optimized input images.
We train the patterns from 500 clean VGG models and 500 poisoned VGG models provided in its official GitHub repository.
Then, we attack five different VGG models on CIFAR-10, and they all can bypass ULP. ULP assumes the trigger is a small patch, while our trigger is not a patch.

\noindent
\textbf{Neural Attention Distillation~\cite{li2021neural}.}
NAD~\cite{li2021neural} is a Trojan removing method.
It first obtains a teacher model by fine-tuning on a set of clean samples.
Then, NAD uses the obtained teacher model to guide the distillation of the Trojan student model to make the intermediate-layer attention of the student model align with that of the teacher model.
To evaluate if our method is resilient to NAD, we conduct experiments on three datasets (i.e., CIFAR-10, GTSRB, and CelebA).
For CIFAR10 and GTSRB, we use Pre-activation ResNet18.
For CelebA, we use ResNet18.
For the implementation of NAD, we use the official code and default hyperparameters specified in the original paper.
In detail, we assume the defender can access 5\% of clean training data.
The initial learning rate is 0.1, and the learning rate is divided by ten after every two epochs.
The data augmentations used are random crop, horizontal flipping, and Cutout~\cite{devries2017improved}.
The results are demonstrated in \autoref{tab:nad}.
For CIFAR-10 and GTSRB, although the ASRs for defended models are low, however, the BAs decrease dramatically after NAD defense.
For CelebA, the defended model still achieves 47.89\% ASR with the BA drop from 79.06\% to 67.52\%.
The results show that our attack is resilient to NAD.

\begin{table}[H]
    \centering
    \scriptsize
    \setlength\tabcolsep{3pt}
    \scalebox{1}{
    \begin{tabular}{@{}cccccccc@{}}
    \toprule
    \multirow{2}{*}{Dataset} &  & \multicolumn{2}{c}{No defense} &  & \multicolumn{2}{c}{NAD} &  \\ \cmidrule(lr){3-4} \cmidrule(lr){6-7}
                             &  & BA             & ASR           &  & BA         & ASR        &  \\ \midrule
    CIFAR-10                 &  & 94.54\%        & 99.91\%       &  & 39.14\%    & 12.07\%    &  \\
    GTSRB                    &  & 99.25\%        & 99.96\%       &  & 14.21\%    & 2.15\%     &  \\
    CelebA                   &  & 79.06\%        & 99.99\%       &  & 67.52\%    & 47.89\%    &  \\ \bottomrule
    \end{tabular}}
    \caption{Resilient to Neural Attention Distillation}
    \label{tab:nad}
\end{table}

\subsection{Compared with ISSBA~\cite{li2021invisible}}\label{sec:issba}
ISSBA~\cite{li2021invisible} is a representative auxiliary models based attacks. 
It first trains an auto-encoder as a Trojan transformation function and then uses it to inject Trojans into victim models.
Following ISSBA~\cite{li2021invisible}, we run our method on a 200 classes subset of ImageNet (specified in Li et al.~\cite{li2021invisible}) and ResNet18 model, and compare our method to it. 
The results are shown in \autoref{tab:imagenet}, where ET means the extra time cost for training the victim model.
Our attack is more efficient with comparable or better ASR and BA, compared with ISSBA. 
The computational and time overhead of our method is much smaller than that of generator/auto-encoder based attacks~\cite{cheng2020deep,li2021invisible,doan2021lira}. 
In detail, the training time of our method is only 19.04\% longer than that of standard training. 
For ImageNet's 200 classes subset, ISSBA~\cite{li2021invisible} takes 7h30mins to train the encoder-decoder. 
However, the extra training time for our method is only 1h18mins on the same dataset.
For stealthiness, 
it is clear that the example of our attack is more close to the original image, while the example of ISSBA has some unnatural ``black fog''. (See Fig.1 in main paper.)

\begin{table}[H]
    \centering
    \scriptsize
    \setlength\tabcolsep{3pt}
    \scalebox{1}{
    \begin{tabular}{@{}ccccccccccc@{}}
    \toprule
    \multirow{2}{*}{Dataset} &  & Non-attack &  & \multicolumn{3}{c}{ISSBA} &  & \multicolumn{3}{c}{BppAttack} \\ \cmidrule(lr){3-3} \cmidrule(lr){5-7} \cmidrule(l){9-11} 
                             &  & BA         &  & BA       & ASR     & ET   &  & BA        & ASR       & ET    \\ \midrule
    ImageNet                 &  & 85.83\%    &  & 85.51\%  & 99.54\% & 450m &  & 85.76\%   & 99.78\%   & 78m   \\ \bottomrule
    \end{tabular}}
    \caption{Effectiveness on ImageNet}\label{tab:imagenet}
\end{table}

\subsection{Compared with WaNet~\cite{nguyen2021wanet}}\label{sec:wanet_our_protocol}
Our method and WaNet~\cite{nguyen2021wanet} have different training protocols. Besides the comparison under different training protocols, we also compare \sys and WaNet under our protocol to further investigate the effectiveness of our proposed quantization triggers.
We compare our method and WaNet under our training protocol on CIFAR-10 and GTSRB. 
The model used is Pre-activation ResNet18 and ResNet18, respectively. 
The results are demonstrated in \autoref{tab:comparison_wanet_same_training}.
Results show that both BA and ASR of our trigger are higher than that of WaNet, showing that the purposed quantization trigger is better than WaNet's trigger.

\begin{table}[H]
    \centering
    \scriptsize
    \setlength\tabcolsep{3pt}
    \scalebox{1}{
    \begin{tabular}{@{}ccccccc@{}}
    \toprule
    \multirow{2}{*}{Dataset} &  & \multicolumn{2}{c}{WaNet} &  & \multicolumn{2}{c}{BppAttack} \\ \cmidrule(lr){3-4} \cmidrule(l){6-7} 
                             &  & BA          & ASR         &  & BA            & ASR           \\ \midrule
    CIFAR-10                 &  & 94.06\%     & 99.35\%     &  & 94.54\%       & 99.91\%       \\
    GTSRB                    &  & 98.45\%     & 98.52\%     &  & 99.25\%       & 99.96\%       \\ \bottomrule
    \end{tabular}}
    \caption{Comparisons to WaNet using our training protocol}
    \label{tab:comparison_wanet_same_training}
    \end{table}

\subsection{Robustness against fine-tuning}
Besides the threat model that assumes the victim users directly deploy the malicious models, here we also consider a transfer learning scenario where the downstream users fine-tune the Trojan model weights with out-of-distribution data.
In some cases, the downstream users even fine-tune the model with different quality of images, and some may incorporate similar quantization techniques to the proposed attack, e.g., JPEG.
Note that injecting Trojans that are robust against fine-tuning is orthogonal to our paper and has been studied by another line of work~\cite{yao2019latent}.
Such approaches can be adopted by us.
By combining with Yao et al.~\cite{yao2019latent}, our attack on CIFAR-10 and ResNet18 can achieve 86.52\% ASR after fine-tuning on 5000 JPEG compressed samples.

\subsection{Discussion: Trojan Triggers}
Traditional Trojan attacks use fixed patterns/noise as Trojan triggers.
Let \(\widetilde{\bm x}\) be the Trojan sample and \(\bm x\) be the corresponding clean sample.
These attacks can be formalized as \(\widetilde{\bm x} = \bm m \odot \bm t + (1-\bm m) \odot \bm x\) (where \(\bm m\) and \(\bm t\) are predefined Trojan trigger mask and pattern) or \(\widetilde{\bm x} = \bm x + \bm \delta\) (where \(\bm \delta\) is the fixed noise).
However, the Trojan triggers are not necessarily a fixed pattern.
Instead, it can be a universal input activity (e.g., quantization, auto-encoder, GAN, or other input transformations), and it can be formalized as \(\widetilde{\bm x} = T(\bm x)\).
The traditional trigger that requires a fixed pattern is actually a special case of the activity function \(T(\bm x)\).

%% file: figtex/appendix_diff_attacks.tex
\begin{figure*}[]
	\centering
	\footnotesize
	\includegraphics[width=1.5\columnwidth]{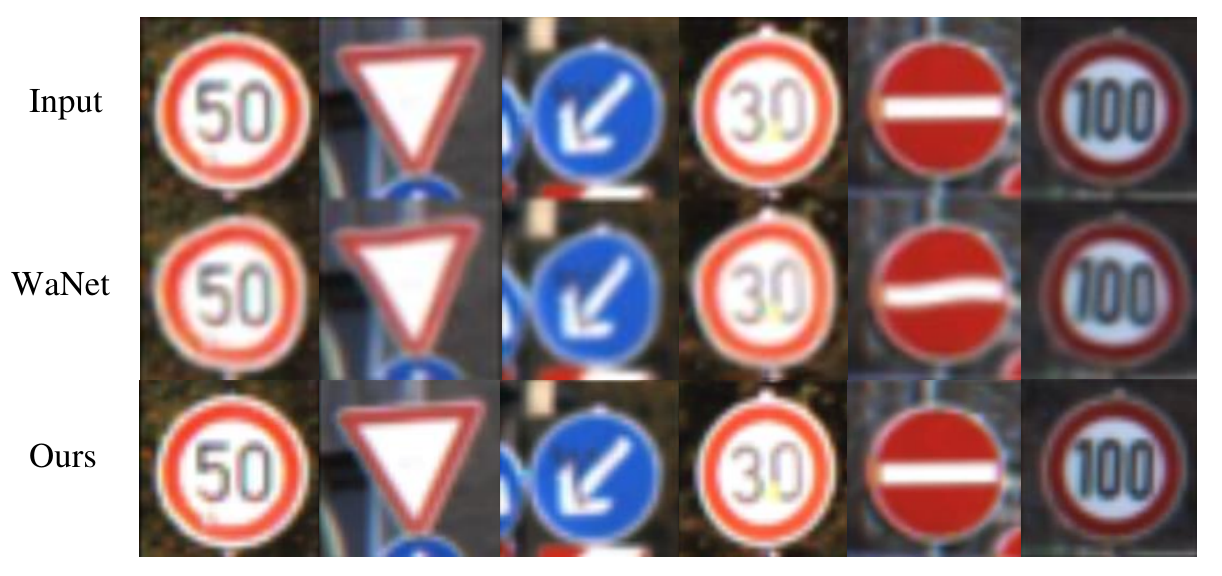}
	\caption{Additional images for comparison between WaNet and our method.}\label{fig:appendix_diff_attacks}
\end{figure*}

%% file: tf/mnist_classfier.tex
\begin{table}[H]
    \centering
    \scriptsize
    \setlength\tabcolsep{4pt}
    \begin{tabular}{@{}cccccc@{}}
        \toprule
        Layer Type & \# of Channels & Filter Size & Stride & Padding & Activation \\ \midrule
        Conv*       & 32             & 3x3         & 2      & 1       & ReLU       \\
        Conv*       & 64             & 3x3         & 2      & 0       & ReLU       \\
        Conv       & 64             & 3x3         & 2      & 0       & ReLU       \\
        FC\dag         & 512            & -           & -      & 0       & ReLU       \\
        FC         & 10             & -           & -      & 0       & Softmax    \\ \bottomrule
    \end{tabular}
    \caption{Details of classifier used for MNIST. FC stands for fully-connected
    layer. * denotes the layer is followed by  a BatchNormalization layer. \dag denotes the layer is followed by  a DropOut layer.}\label{tab:mnist_classifier}
    \end{table}

%% file: figtex/appendix_diff_bits.tex
\begin{figure*}[]
	\centering
	\footnotesize
	\includegraphics[width=2\columnwidth]{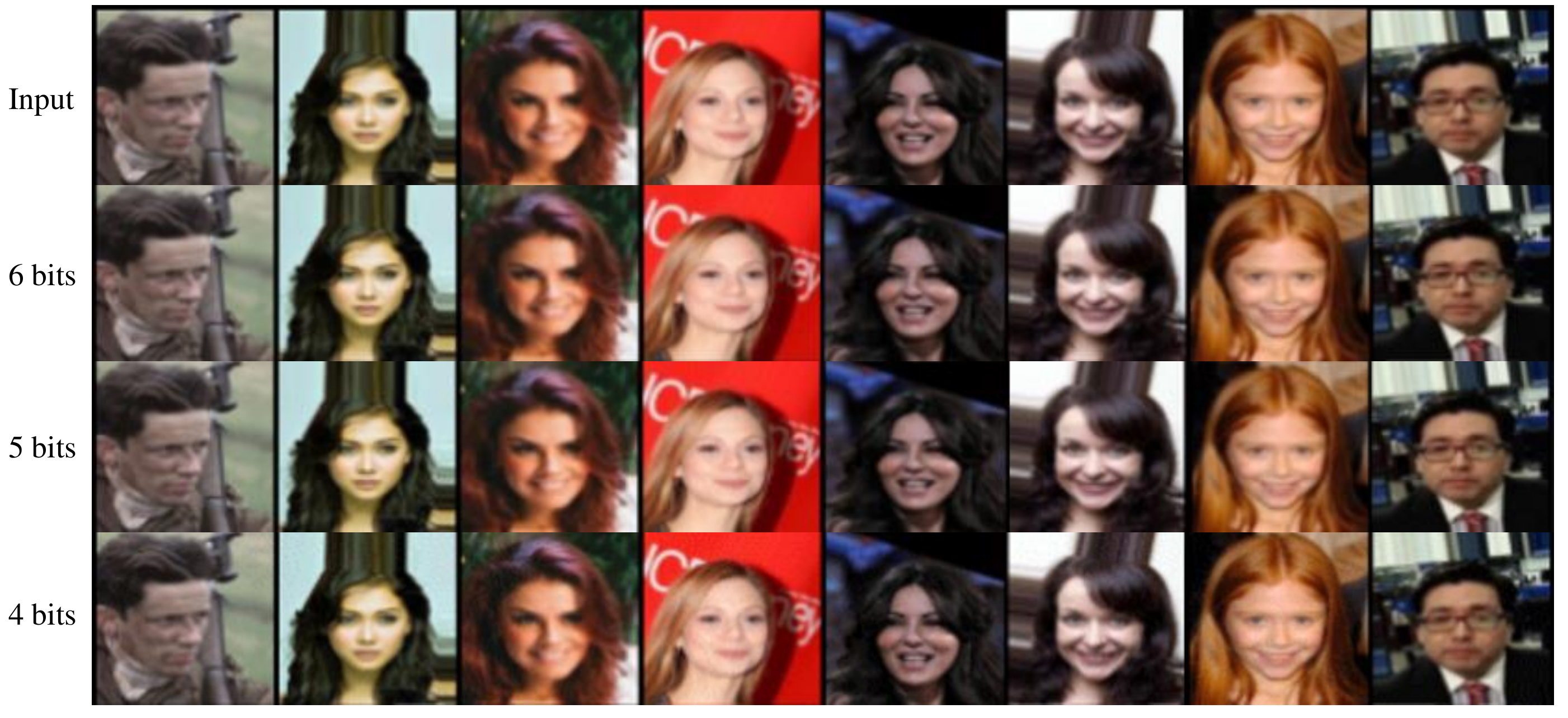}
	\caption{Additional images to demonstrate the influence of different bits numbers.}\label{fig:appendix_diff_bits}
\end{figure*}

%% file: figtex/ss.tex
\begin{figure}[H]
	\centering
	\footnotesize
	\includegraphics[width=0.7\columnwidth]{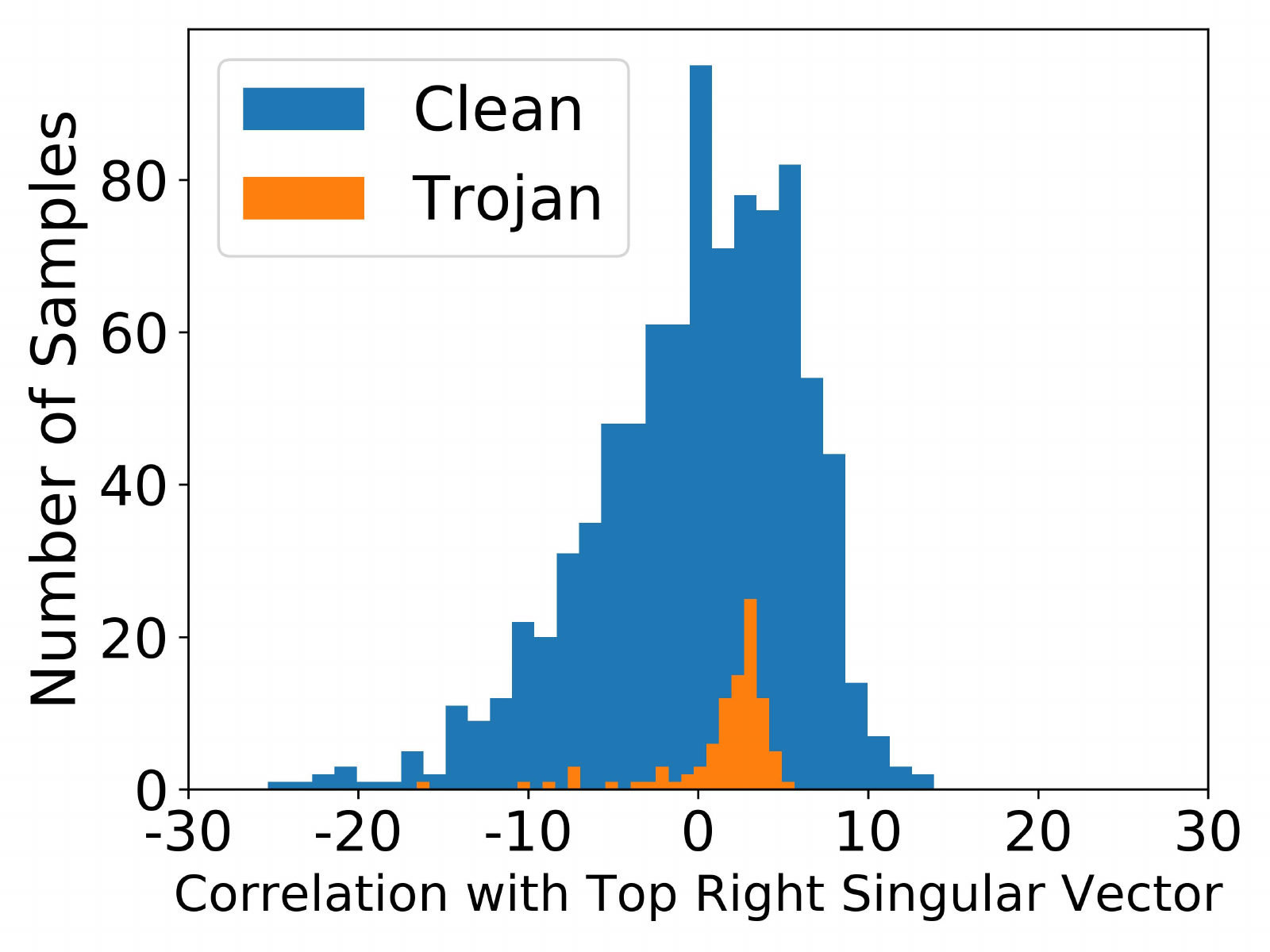}
	\caption{Resilient to Spectral Signature.}\label{fig:ss}
\end{figure}

%% file: main.bbl
\begin{thebibliography}{10}\itemsep=-1pt

\bibitem{TrojAI:online}
Trojai.
\newblock \url{https://pages.nist.gov/trojai/docs/about.html/}.

\bibitem{barni2019new}
Mauro Barni, Kassem Kallas, and Benedetta Tondi.
\newblock A new backdoor attack in cnns by training set corruption without
  label poisoning.
\newblock In {\em 2019 IEEE International Conference on Image Processing
  (ICIP)}, pages 101--105. IEEE, 2019.

\bibitem{bloomberg2008color}
Dan~S Bloomberg.
\newblock Color quantization using octrees.
\newblock {\em Leptonica, ss}, pages 1--10, 2008.

\bibitem{carlini2021poisoning}
Nicholas Carlini and Andreas Terzis.
\newblock Poisoning and backdooring contrastive learning.
\newblock {\em arXiv preprint arXiv:2106.09667}, 2021.

\bibitem{celebi2011improving}
M~Emre Celebi.
\newblock Improving the performance of k-means for color quantization.
\newblock {\em Image and Vision Computing}, 29(4):260--271, 2011.

\bibitem{chen2018detecting}
Bryant Chen, Wilka Carvalho, Nathalie Baracaldo, Heiko Ludwig, Benjamin
  Edwards, Taesung Lee, Ian Molloy, and Biplav Srivastava.
\newblock Detecting backdoor attacks on deep neural networks by activation
  clustering.
\newblock {\em SafeAI@AAAI}, 2019.

\bibitem{chen2019deepinspect}
Huili Chen, Cheng Fu, Jishen Zhao, and Farinaz Koushanfar.
\newblock Deepinspect: A black-box trojan detection and mitigation framework
  for deep neural networks.
\newblock In {\em IJCAI}, pages 4658--4664, 2019.

\bibitem{chen2017deeplab}
Liang-Chieh Chen, George Papandreou, Iasonas Kokkinos, Kevin Murphy, and Alan~L
  Yuille.
\newblock Deeplab: Semantic image segmentation with deep convolutional nets,
  atrous convolution, and fully connected crfs.
\newblock {\em IEEE transactions on pattern analysis and machine intelligence},
  40(4):834--848, 2017.

\bibitem{chen2017targeted}
Xinyun Chen, Chang Liu, Bo Li, Kimberly Lu, and Dawn Song.
\newblock Targeted backdoor attacks on deep learning systems using data
  poisoning.
\newblock {\em arXiv preprint arXiv:1712.05526}, 2017.

\bibitem{cheng2020deep}
Siyuan Cheng, Yingqi Liu, Shiqing Ma, and Xiangyu Zhang.
\newblock Deep feature space trojan attack of neural networks by controlled
  detoxification.
\newblock {\em AAAI}, 2021.

\bibitem{chou2018sentinet}
Edward Chou, Florian Tram{\`e}r, Giancarlo Pellegrino, and Dan Boneh.
\newblock Sentinet: Detecting physical attacks against deep learning systems.
\newblock 2018.

\bibitem{devries2017improved}
Terrance DeVries and Graham~W Taylor.
\newblock Improved regularization of convolutional neural networks with cutout.
\newblock {\em arXiv preprint arXiv:1708.04552}, 2017.

\bibitem{doan2020februus}
Bao~Gia Doan, Ehsan Abbasnejad, and Damith~C Ranasinghe.
\newblock Februus: Input purification defense against trojan attacks on deep
  neural network systems.
\newblock In {\em Annual Computer Security Applications Conference}, pages
  897--912, 2020.

\bibitem{doan2021lira}
Khoa Doan, Yingjie Lao, Weijie Zhao, and Ping Li.
\newblock Lira: Learnable, imperceptible and robust backdoor attacks.
\newblock In {\em Proceedings of the IEEE/CVF International Conference on
  Computer Vision}, pages 11966--11976, 2021.

\bibitem{du2019robust}
Min Du, Ruoxi Jia, and Dawn Song.
\newblock Robust anomaly detection and backdoor attack detection via
  differential privacy.
\newblock {\em International Conference on Learning Representations (ICLR)},
  2020.

\bibitem{Floyd:1976:AAS}
Robert~W. Floyd and Louis Steinberg.
\newblock {A}n {A}daptive {A}lgorithm for {S}patial {G}reyscale.
\newblock {\em Proceedings of the Society for Information Display},
  17(2):75--77, 1976.

\bibitem{gao2019strip}
Yansong Gao, Change Xu, Derui Wang, Shiping Chen, Damith~C Ranasinghe, and
  Surya Nepal.
\newblock Strip: A defence against trojan attacks on deep neural networks.
\newblock In {\em Proceedings of the 35th Annual Computer Security Applications
  Conference}, pages 113--125, 2019.

\bibitem{goodfellow2014explaining}
Ian~J Goodfellow, Jonathon Shlens, and Christian Szegedy.
\newblock Explaining and harnessing adversarial examples.
\newblock {\em International Conference on Learning Representations (ICLR)},
  2015.

\bibitem{gu2017badnets}
Tianyu Gu, Brendan Dolan-Gavitt, and Siddharth Garg.
\newblock Badnets: Identifying vulnerabilities in the machine learning model
  supply chain.
\newblock {\em arXiv preprint arXiv:1708.06733}, 2017.

\bibitem{hampel1974influence}
Frank~R Hampel.
\newblock The influence curve and its role in robust estimation.
\newblock {\em Journal of the american statistical association},
  69(346):383--393, 1974.

\bibitem{hayase2021spectre}
Jonathan Hayase, Weihao Kong, Raghav Somani, and Sewoong Oh.
\newblock Spectre: Defending against backdoor attacks using robust statistics.
\newblock {\em International Conference on Machine Learning}, 2021.

\bibitem{he2016deep}
Kaiming He, Xiangyu Zhang, Shaoqing Ren, and Jian Sun.
\newblock Deep residual learning for image recognition.
\newblock In {\em Proceedings of the IEEE conference on computer vision and
  pattern recognition}, pages 770--778, 2016.

\bibitem{he2016identity}
Kaiming He, Xiangyu Zhang, Shaoqing Ren, and Jian Sun.
\newblock Identity mappings in deep residual networks.
\newblock In {\em European conference on computer vision}, pages 630--645.
  Springer, 2016.

\bibitem{heckbert1982color}
Paul Heckbert.
\newblock Color image quantization for frame buffer display.
\newblock {\em ACM Siggraph Computer Graphics}, 16(3):297--307, 1982.

\bibitem{hong2020effectiveness}
Sanghyun Hong, Varun Chandrasekaran, Yi{\u{g}}itcan Kaya, Tudor Dumitra{\c{s}},
  and Nicolas Papernot.
\newblock On the effectiveness of mitigating data poisoning attacks with
  gradient shaping.
\newblock {\em arXiv preprint arXiv:2002.11497}, 2020.

\bibitem{hu2018squeeze}
Jie Hu, Li Shen, and Gang Sun.
\newblock Squeeze-and-excitation networks.
\newblock In {\em Proceedings of the IEEE conference on computer vision and
  pattern recognition}, pages 7132--7141, 2018.

\bibitem{hu2016simple}
Xiangyu~Y Hu.
\newblock Simple gradient-based error-diffusion method.
\newblock {\em Journal of Electronic Imaging}, 25(4):043029, 2016.

\bibitem{huang2017densely}
Gao Huang, Zhuang Liu, Laurens Van Der~Maaten, and Kilian~Q Weinberger.
\newblock Densely connected convolutional networks.
\newblock In {\em Proceedings of the IEEE conference on computer vision and
  pattern recognition}, pages 4700--4708, 2017.

\bibitem{jacobs1991retinal}
Gerald~H Jacobs, Jay Neitz, and Jess~F Deegan.
\newblock Retinal receptors in rodents maximally sensitive to ultraviolet
  light.
\newblock {\em Nature}, 353(6345):655--656, 1991.

\bibitem{jia2021badencoder}
Jinyuan Jia, Yupei Liu, and Neil~Zhenqiang Gong.
\newblock Badencoder: Backdoor attacks to pre-trained encoders in
  self-supervised learning.
\newblock {\em 2022 IEEE Symposium on Security and Privacy (SP). IEEE}, 2022.

\bibitem{judd1952color}
Deane~B Judd.
\newblock Color in business, science and industry.
\newblock 1952.

\bibitem{khosla2020supervised}
Prannay Khosla, Piotr Teterwak, Chen Wang, Aaron Sarna, Yonglong Tian, Phillip
  Isola, Aaron Maschinot, Ce Liu, and Dilip Krishnan.
\newblock Supervised contrastive learning.
\newblock {\em arXiv preprint arXiv:2004.11362}, 2020.

\bibitem{kolouri2020universal}
Soheil Kolouri, Aniruddha Saha, Hamed Pirsiavash, and Heiko Hoffmann.
\newblock Universal litmus patterns: Revealing backdoor attacks in cnns.
\newblock In {\em Proceedings of the IEEE/CVF Conference on Computer Vision and
  Pattern Recognition}, pages 301--310, 2020.

\bibitem{krizhevsky2009learning}
Alex Krizhevsky, Geoffrey Hinton, et~al.
\newblock Learning multiple layers of features from tiny images.
\newblock 2009.

\bibitem{lecun1998gradient}
Yann LeCun, L{\'e}on Bottou, Yoshua Bengio, and Patrick Haffner.
\newblock Gradient-based learning applied to document recognition.
\newblock {\em Proceedings of the IEEE}, 86(11):2278--2324, 1998.

\bibitem{li2020deep}
Shaofeng Li, Shiqing Ma, Minhui Xue, and Benjamin Zi~Hao Zhao.
\newblock Deep learning backdoors.
\newblock {\em arXiv preprint arXiv:2007.08273}, 2020.

\bibitem{li2021neural}
Yige Li, Nodens Koren, Lingjuan Lyu, Xixiang Lyu, Bo Li, and Xingjun Ma.
\newblock Neural attention distillation: Erasing backdoor triggers from deep
  neural networks.
\newblock {\em International Conference on Learning Representations (ICLR)},
  2021.

\bibitem{li2021invisible}
Yuezun Li, Yiming Li, Baoyuan Wu, Longkang Li, Ran He, and Siwei Lyu.
\newblock Invisible backdoor attack with sample-specific triggers.
\newblock In {\em Proceedings of the IEEE/CVF International Conference on
  Computer Vision}, pages 16463--16472, 2021.

\bibitem{lin2020composite}
Junyu Lin, Lei Xu, Yingqi Liu, and Xiangyu Zhang.
\newblock Composite backdoor attack for deep neural network by mixing existing
  benign features.
\newblock In {\em Proceedings of the 2020 ACM SIGSAC Conference on Computer and
  Communications Security}, pages 113--131, 2020.

\bibitem{liu2018fine}
Kang Liu, Brendan Dolan-Gavitt, and Siddharth Garg.
\newblock Fine-pruning: Defending against backdooring attacks on deep neural
  networks.
\newblock In {\em International Symposium on Research in Attacks, Intrusions,
  and Defenses}, pages 273--294. Springer, 2018.

\bibitem{liu2019abs}
Yingqi Liu, Wen-Chuan Lee, Guanhong Tao, Shiqing Ma, Yousra Aafer, and Xiangyu
  Zhang.
\newblock Abs: Scanning neural networks for back-doors by artificial brain
  stimulation.
\newblock In {\em Proceedings of the 2019 ACM SIGSAC Conference on Computer and
  Communications Security}, pages 1265--1282, 2019.

\bibitem{liu2017trojaning}
Yingqi Liu, Shiqing Ma, Yousra Aafer, Wen-Chuan Lee, Juan Zhai, Weihang Wang,
  and Xiangyu Zhang.
\newblock Trojaning attack on neural networks.
\newblock {\em NDSS}, 2018.

\bibitem{liu2020reflection}
Yunfei Liu, Xingjun Ma, James Bailey, and Feng Lu.
\newblock Reflection backdoor: A natural backdoor attack on deep neural
  networks.
\newblock In {\em European Conference on Computer Vision}, pages 182--199.
  Springer, 2020.

\bibitem{liu2021ex}
Yingqi Liu, Guangyu Shen, Guanhong Tao, Zhenting Wang, Shiqing Ma, and Xiangyu
  Zhang.
\newblock Ex-ray: Distinguishing injected backdoor from natural features in
  neural networks by examining differential feature symmetry.
\newblock {\em arXiv preprint arXiv:2103.08820}, 2021.

\bibitem{liu2015faceattributes}
Ziwei Liu, Ping Luo, Xiaogang Wang, and Xiaoou Tang.
\newblock Deep learning face attributes in the wild.
\newblock In {\em Proceedings of International Conference on Computer Vision
  (ICCV)}, December 2015.

\bibitem{moosavi2016deepfool}
Seyed-Mohsen Moosavi-Dezfooli, Alhussein Fawzi, and Pascal Frossard.
\newblock Deepfool: a simple and accurate method to fool deep neural networks.
\newblock In {\em Proceedings of the IEEE conference on computer vision and
  pattern recognition}, pages 2574--2582, 2016.

\bibitem{nadenau2000human}
Marcus~J Nadenau, Stefan Winkler, David Alleysson, and Murat Kunt.
\newblock Human vision models for perceptually optimized image processing--a
  review.
\newblock {\em Proceedings of the IEEE}, 32, 2000.

\bibitem{neitz1986polymorphism}
Jay Neitz and Gerald~H Jacobs.
\newblock Polymorphism of the long-wavelength cone in normal human colour
  vision.
\newblock {\em Nature}, 323(6089):623--625, 1986.

\bibitem{nguyen2021wanet}
Anh Nguyen and Anh Tran.
\newblock Wanet--imperceptible warping-based backdoor attack.
\newblock {\em arXiv preprint arXiv:2102.10369}, 2021.

\bibitem{nguyen2020input}
Tuan~Anh Nguyen and Anh Tran.
\newblock Input-aware dynamic backdoor attack.
\newblock {\em Advances in Neural Information Processing Systems},
  33:3454--3464, 2020.

\bibitem{orekondy2019knockoff}
Tribhuvanesh Orekondy, Bernt Schiele, and Mario Fritz.
\newblock Knockoff nets: Stealing functionality of black-box models.
\newblock In {\em Proceedings of the IEEE/CVF Conference on Computer Vision and
  Pattern Recognition}, pages 4954--4963, 2019.

\bibitem{pambrun2015limitations}
Jean-Fran{\c{c}}ois Pambrun and Rita Noumeir.
\newblock Limitations of the ssim quality metric in the context of diagnostic
  imaging.
\newblock In {\em 2015 IEEE International Conference on Image Processing
  (ICIP)}, pages 2960--2963. IEEE, 2015.

\bibitem{ren2015faster}
Shaoqing Ren, Kaiming He, Ross Girshick, and Jian Sun.
\newblock Faster r-cnn: Towards real-time object detection with region proposal
  networks.
\newblock {\em Advances in neural information processing systems}, 28:91--99,
  2015.

\bibitem{salem2020dynamic}
Ahmed Salem, Rui Wen, Michael Backes, Shiqing Ma, and Yang Zhang.
\newblock Dynamic backdoor attacks against machine learning models.
\newblock {\em IEEE European Symposium on Security and Privacy (EuroS\&P)},
  2022.

\bibitem{salem2018ml}
Ahmed Salem, Yang Zhang, Mathias Humbert, Pascal Berrang, Mario Fritz, and
  Michael Backes.
\newblock Ml-leaks: Model and data independent membership inference attacks and
  defenses on machine learning models.
\newblock {\em NDSS}, 2019.

\bibitem{sandler2018mobilenetv2}
Mark Sandler, Andrew Howard, Menglong Zhu, Andrey Zhmoginov, and Liang-Chieh
  Chen.
\newblock Mobilenetv2: Inverted residuals and linear bottlenecks.
\newblock In {\em Proceedings of the IEEE conference on computer vision and
  pattern recognition}, pages 4510--4520, 2018.

\bibitem{selvaraju2017grad}
Ramprasaath~R Selvaraju, Michael Cogswell, Abhishek Das, Ramakrishna Vedantam,
  Devi Parikh, and Dhruv Batra.
\newblock Grad-cam: Visual explanations from deep networks via gradient-based
  localization.
\newblock In {\em Proceedings of the IEEE international conference on computer
  vision}, pages 618--626, 2017.

\bibitem{shen2021backdoor}
Guangyu Shen, Yingqi Liu, Guanhong Tao, Shengwei An, Qiuling Xu, Siyuan Cheng,
  Shiqing Ma, and Xiangyu Zhang.
\newblock Backdoor scanning for deep neural networks through k-arm
  optimization.
\newblock In {\em International Conference on Machine Learning}, pages
  9525--9536. PMLR, 2021.

\bibitem{shokri2017membership}
Reza Shokri, Marco Stronati, Congzheng Song, and Vitaly Shmatikov.
\newblock Membership inference attacks against machine learning models.
\newblock In {\em 2017 IEEE Symposium on Security and Privacy (SP)}, pages
  3--18. IEEE, 2017.

\bibitem{stallkamp2012man}
Johannes Stallkamp, Marc Schlipsing, Jan Salmen, and Christian Igel.
\newblock Man vs. computer: Benchmarking machine learning algorithms for
  traffic sign recognition.
\newblock {\em Neural networks}, 32:323--332, 2012.

\bibitem{tao2022model}
Guanhong Tao, Yingqi Liu, Guangyu Shen, Qiuling Xu, Shengwei An, Zhuo Zhang,
  and Xiangyu Zhang.
\newblock Model orthogonalization: Class distance hardening in neural networks
  for better security.
\newblock In {\em 2022 IEEE Symposium on Security and Privacy (SP). IEEE},
  2022.

\bibitem{tran2018spectral}
Brandon Tran, Jerry Li, and Aleksander Madry.
\newblock Spectral signatures in backdoor attacks.
\newblock {\em Advances in Neural Information Processing Systems}, 2018.

\bibitem{truong2021data}
Jean-Baptiste Truong, Pratyush Maini, Robert~J Walls, and Nicolas Papernot.
\newblock Data-free model extraction.
\newblock In {\em Proceedings of the IEEE/CVF Conference on Computer Vision and
  Pattern Recognition}, pages 4771--4780, 2021.

\bibitem{ulichney1993void}
Robert~A Ulichney.
\newblock Void-and-cluster method for dither array generation.
\newblock In {\em Human Vision, Visual Processing, and Digital Display IV},
  volume 1913, pages 332--343. International Society for Optics and Photonics,
  1993.

\bibitem{veldanda2020nnoculation}
Akshaj~Kumar Veldanda, Kang Liu, Benjamin Tan, Prashanth Krishnamurthy, Farshad
  Khorrami, Ramesh Karri, Brendan Dolan-Gavitt, and Siddharth Garg.
\newblock Nnoculation: broad spectrum and targeted treatment of backdoored
  dnns.
\newblock {\em arXiv preprint arXiv:2002.08313}, 2020.

\bibitem{verevka1995local}
Oleg Verevka.
\newblock The local k-means algorithm for colour image quantization.
\newblock 1995.

\bibitem{wang2019neural}
Bolun Wang, Yuanshun Yao, Shawn Shan, Huiying Li, Bimal Viswanath, Haitao
  Zheng, and Ben~Y Zhao.
\newblock Neural cleanse: Identifying and mitigating backdoor attacks in neural
  networks.
\newblock In {\em 2019 IEEE Symposium on Security and Privacy (SP)}, pages
  707--723. IEEE, 2019.

\bibitem{wang2022towards}
Zhenting Wang, Hailun Ding, Juan Zhai, and Shiqing Ma.
\newblock Towards understanding and defending input space trojans.
\newblock {\em arXiv preprint arXiv:2202.06382}, 2022.

\bibitem{xie2019dba}
Chulin Xie, Keli Huang, Pin-Yu Chen, and Bo Li.
\newblock Dba: Distributed backdoor attacks against federated learning.
\newblock In {\em International Conference on Learning Representations (ICLR)},
  2019.

\bibitem{xie2017aggregated}
Saining Xie, Ross Girshick, Piotr Doll{\'a}r, Zhuowen Tu, and Kaiming He.
\newblock Aggregated residual transformations for deep neural networks.
\newblock In {\em Proceedings of the IEEE conference on computer vision and
  pattern recognition}, pages 1492--1500, 2017.

\bibitem{xu2017feature}
Weilin Xu, David Evans, and Yanjun Qi.
\newblock Feature squeezing: Detecting adversarial examples in deep neural
  networks.
\newblock {\em NDSS}, 2018.

\bibitem{yao2019latent}
Yuanshun Yao, Huiying Li, Haitao Zheng, and Ben~Y Zhao.
\newblock Latent backdoor attacks on deep neural networks.
\newblock In {\em Proceedings of the 2019 ACM SIGSAC Conference on Computer and
  Communications Security}, pages 2041--2055, 2019.

\bibitem{zeki1980representation}
Semir Zeki.
\newblock The representation of colours in the cerebral cortex.
\newblock {\em Nature}, 284(5755):412--418, 1980.

\bibitem{zhao2020bridging}
Pu Zhao, Pin-Yu Chen, Payel Das, Karthikeyan~Natesan Ramamurthy, and Xue Lin.
\newblock Bridging mode connectivity in loss landscapes and adversarial
  robustness.
\newblock {\em International Conference on Learning Representations (ICLR)},
  2020.

\bibitem{zhu2017unpaired}
Jun-Yan Zhu, Taesung Park, Phillip Isola, and Alexei~A Efros.
\newblock Unpaired image-to-image translation using cycle-consistent
  adversarial networks.
\newblock In {\em Proceedings of the IEEE international conference on computer
  vision}, pages 2223--2232, 2017.

\end{thebibliography}
